%% file: zoomnet-arxiv.tex
\documentclass[10pt,twocolumn,letterpaper]{article}

\usepackage[pagenumbers]{cvpr} % To force page numbers, e.g. for an arXiv version

\usepackage{graphicx}
\usepackage{amsmath}
\usepackage{amssymb}
\usepackage{booktabs}

\usepackage{url}
\usepackage{enumitem}
\usepackage{multirow}
\usepackage{colortbl}
\definecolor{tabtitle}{gray}{.8}
\definecolor{ours}{gray}{.95}
\definecolor{ggray}{RGB}{127,127,127}
\definecolor{reda}{RGB}{202,0,0}
\definecolor{redb}{RGB}{217,148,143}
\definecolor{myyellow}{RGB}{190,144,0}
\definecolor{mygreen}{RGB}{0,136,51}
\definecolor{myblue}{RGB}{0,102,204}
\usepackage{pifont}
\newcommand{\yes}{\text{\ding{51}}}
\newcommand{\no}{\text{\ding{55}}}
\newcommand{\none}{—}

\usepackage{listings}
\definecolor{codegreen}{rgb}{0,0.6,0}
\definecolor{codegray}{rgb}{0.5,0.5,0.5}
\definecolor{codepurple}{rgb}{0.58,0,0.82}
\definecolor{backcolour}{rgb}{0.95,0.95,0.92}
\lstdefinestyle{mystyle}{
    backgroundcolor=\color{backcolour},
    commentstyle=\color{codegreen},
    keywordstyle=\color{magenta},
    numberstyle=\tiny\color{codegray},
    stringstyle=\color{codepurple},
    basicstyle=\ttfamily\footnotesize,
    breakatwhitespace=false,
    breaklines=true,
    captionpos=b,
    keepspaces=true,
    numbers=left,
    numbersep=5pt,
    showspaces=false,
    showstringspaces=false,
    showtabs=false,
    tabsize=2
}
\usepackage{algorithm}
\usepackage{algpseudocode}
\usepackage[pagebackref,breaklinks,colorlinks]{hyperref}

\usepackage[capitalize]{cleveref}
\crefname{section}{Sec.}{Secs.}
\Crefname{section}{Section}{Sections}
\Crefname{table}{Table}{Tables}
\crefname{table}{Tab.}{Tabs.}

\def\cvprPaperID{***} % *** Enter the CVPR Paper ID here
\def\confName{CVPR}
\def\confYear{2022}

\begin{document}

\title{Zoom In and Out: A Mixed-scale Triplet Network \\ for Camouflaged Object Detection}

\author{
Youwei Pang$^1$\protect\footnotemark[2]\,,
Xiaoqi Zhao$^1$\protect\footnotemark[2]\,,
Tian-Zhu Xiang$^3$\,,
Lihe Zhang$^1$\protect\footnotemark[1]\,
and Huchuan Lu$^{1,2}$\\
$^1$Dalian University of Technology, China \quad
$^2$Peng Cheng Laboratory, China \\
$^3$Inception Institute of Artificial Intelligence, UAE \\
{\tt\small
\{lartpang, zxq\}@mail.dlut.edu.cn,
tianzhu.xiang19@gmail.com,
\{zhanglihe, lhchuan\}@dlut.edu.cn}
}

\maketitle
\renewcommand{\thefootnote}{\fnsymbol{footnote}} %将脚注符号设置为fnsymbol类型，即特殊符号表示
\footnotetext[2]{These authors contributed equally to this work.} %对应脚注[1]
\footnotetext[1]{Corresponding author.} %对应脚注[2]
\renewcommand{\thefootnote}{\arabic{footnote}}

%%%%%%%%% ABSTRACT
\begin{abstract}
	The recently proposed camouflaged object detection (COD) attempts to segment objects that are visually blended into their surroundings, which is extremely complex and difficult in real-world scenarios.
	Apart from high intrinsic similarity between the camouflaged objects and their background, the objects are usually diverse in scale, fuzzy in appearance, and even severely occluded.
	To deal with these problems, we propose a mixed-scale triplet network, \textbf{ZoomNet}, which mimics the behavior of humans when observing vague images, i.e., zooming in and out.
	Specifically, our ZoomNet employs the zoom strategy to learn the discriminative mixed-scale semantics by the designed scale integration unit and hierarchical mixed-scale unit, which fully explores imperceptible clues between the candidate objects and background surroundings.
	Moreover, considering the uncertainty and ambiguity derived from indistinguishable textures, we construct a simple yet effective regularization constraint, uncertainty-aware loss, to promote the model to accurately produce predictions with higher confidence in candidate regions.
	Without bells and whistles, our proposed highly task-friendly model consistently surpasses the existing 23 state-of-the-art methods on four public datasets.
	Besides, the superior performance over the recent cutting-edge models on the SOD task also verifies the effectiveness and generality of our model.
	The code will be available at \url{https://github.com/lartpang/ZoomNet}.
\end{abstract}

%%%%%%%%% BODY TEXT
\section{Introduction}

\begin{figure}[t]
	\centering
	\includegraphics[width=\linewidth]{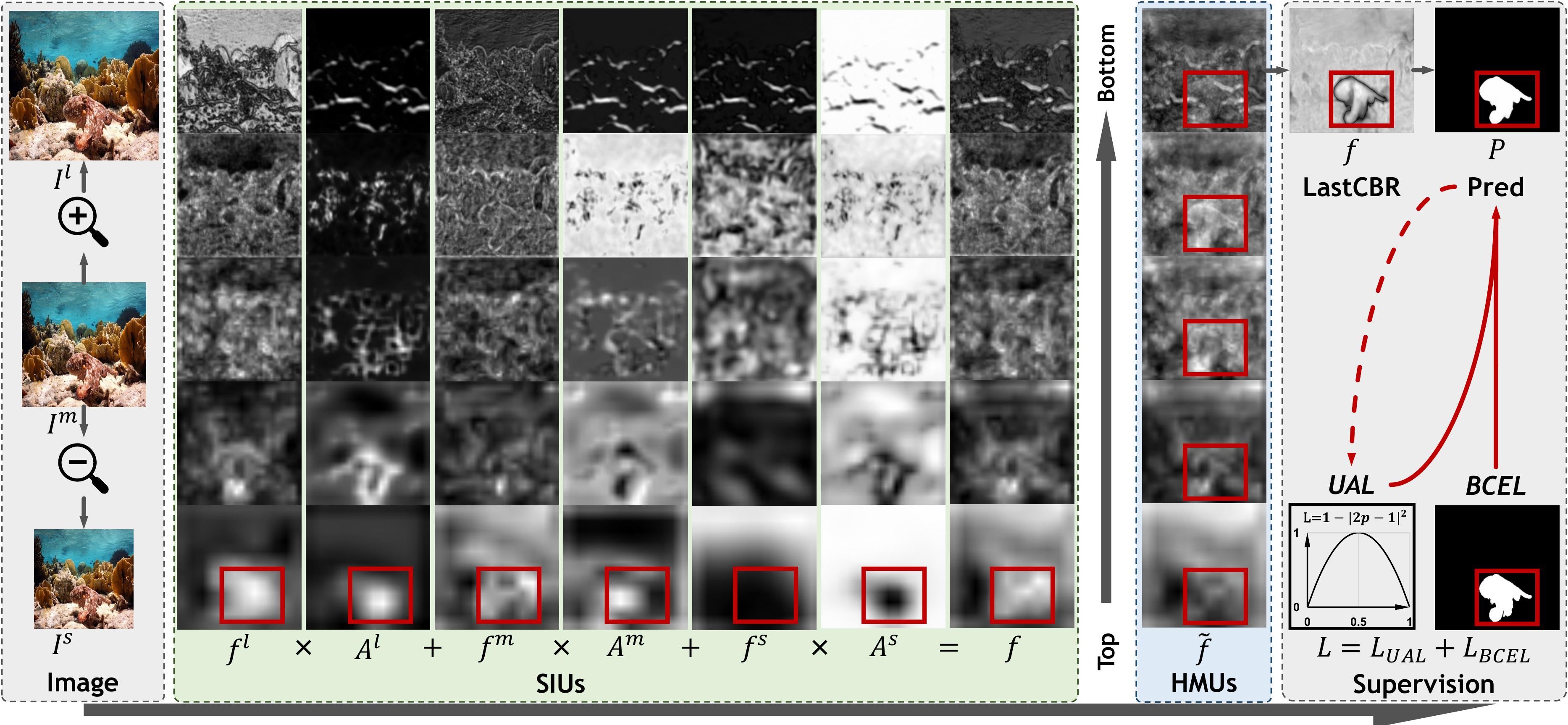}
	\caption{
		Illustration of ZoomNet.
		Based on \textit{zoom} strategy, our model distills the \textit{differentiated} features at different “zoom” scales.
		Then we design SIUs to screen and aggregate scale-specific features, and HMUs to reorganize and enhance mixed-scale features.
		Under the supervision of BCEL and the proposed UAL, the model produces the accurate and reliable camouflaged object prediction.
		Note that BCEL is computed based on ground truth while UAL is not.
		$f$: feature map; $A$: attention map.
		LastCBR: the last ``Conv-BN-ReLU'' layer before the prediction.
		$l/m/s$: Different input scales.
		The whiter region denotes the larger activation response.
	}
	\label{fig:top_img}
\end{figure}

Camouflaged objects are often ``seamlessly'' integrated into the environment by changing their appearance, coloration or pattern to avoid detection, such as chameleons, cuttlefishes and flatfishes.
This is mainly due to their self-protection mechanism in the harsh living environment.
Broadly speaking, camouflaged objects also refer to the objects that are extremely small in size, highly similar to the background, or heavily obscured.
They subtly hide themselves in the surroundings, making them difficult to be found, \textit{e.g.}, soldiers wearing camouflaged uniforms and lions hiding in the grass.
Camouflaged object detection (COD) is far more complex and challenging than traditional salient object detection or other object segmentation.
Recently, it has attracted ever-growing research interest from the computer vision community and facilitates many valuable real-life applications, such as search and rescue~\cite{ConcealedObjectDetection}, species discovery~\cite{EarlyWEvolutionandEcology}, and medical image analysis~\cite{PraNet,Inf-Net,MSNet}.

Recently, numerous deep learning-based methods have been proposed and achieved significant progress.
Nevertheless, they are still struggled to accurately and reliably detect camouflaged objects, due to visual insignificance of camouflaged objects, and high diversity in scale, appearance and occlusion.
By observing our experiments, it is found that the current COD detectors are susceptible to distractors from background surroundings.
Thus it is difficult to excavate discriminative and subtle semantic cues for camouflaged objects, resulting in the inability to clearly segment the camouflaged objects from the chaotic background and the predictions of some uncertain (low-confidence) regions.
Taking these into mind, in this paper, we summarize the COD issue into two aspects:
1) \textit{How to accurately locate camouflaged objects under conditions of inconspicuous appearance and various scales?}
2) \textit{How to suppress the obvious interference from the background and infer camouflaged objects more reliably?}
Intuitively, to accurately find the vague or camouflaged objects in the scene, humans may try to refer to and compare the changes in the shape or appearance at different scales by zooming in and out (re-scaling) the image. This specific behavior pattern of human beings motivates us to identify camouflaged objects by mimicking the zooming in and out strategy.

With this inspiration, in this paper, we propose a mixed-scale triplet network, \textit{ZoomNet}, which significantly improves the existing camouflaged object detection performance.
\textbf{Firstly}, for accurate object location, we employ scale space theory~\cite{ScaleSpaceTheory1,ScaleSpaceTheory2,ScaleSpaceFiltering} to imitate zooming in and out strategy.
Specifically, we design two key modules, \textit{i.e.}, the scale integration unit (SIU) and the hierarchical mixed-scale unit (HMU).
As shown in Fig.~\ref{fig:top_img}, our model extracts differentiated camouflaged object features at different ``zoom'' scales using the triplet architecture, then adopts SIUs to screen and aggregate scale-specific features, and utilizes HMUs to further reorganize and enhance mixed-scale features.
Thus, our model is able to mine the accurate and subtle semantic clues between objects and background under the mixed scales, and produce accurate predictions.
Besides, we use the shared weight strategy, which achieves a good balance of efficiency and effectiveness.
\textbf{Secondly}, it is related to reliable prediction in complex scenarios.
Although the object is accurately located, the indistinguishable texture and background will easily bring negative effects to the model learning, \textit{e.g.} predicting uncertain/ambiguity regions, which greatly reduces the detection performance and cannot be ignored.
This can be seen in Fig.~\ref{fig:visualation} (Row 3 and 4) and Fig. 1 in the \textit{supp}.
To this end, we design an uncertainty-aware loss (UAL) to guide the model training, which is only based on the prior knowledge that a good COD prediction should have a clear polarization trend.
Its GT-independent characteristic makes it suitable for enhancing the GT-based BCE loss.
This targeted enhancement strategy can force the network to optimize the prediction of the uncertain regions during the training process, enabling our ZoomNet to distinguish the uncertain regions and segment the camouflaged objects reliably.

Our contributions can be summarized as follows:
1) For the COD task, we propose a mixed-scale triplet network, \textit{ZoomNet}, which can credibly capture the objects in complex scenes by characterizing and unifying the scale-specific appearance features at different ``zoom'' scales and the purposeful optimization strategy. % in a targeted manner
2) To obtain the discriminative feature representation of camouflaged objects, we design SIUs and HMUs to distill, aggregate and strengthen the scale-specific and subtle semantic representation for accurate COD.
3) We propose a simple yet effective optimization enhancement strategy, UAL, which can significantly suppress the uncertainty and interference from the background without increasing extra parameters.
4) Our model greatly surpasses recent 23 state-of-the-art methods under seven metrics on four COD datasets. Furthermore, it shows good generalization in the SOD task and the superior performance compared with the existing SOD methods.

\section{Related Work}

\noindent{\textbf{Camouflaged Object.}}
The study of camouflage has a long history in biology.
This behavior of creatures in nature can be regarded as the result of natural selection and adaptation.
In fact, in human life and other parts of society, it also has a profound impact, \textit{e.g.}, arts, popular culture, and design.
More details can be found in~\cite{StevensAnimalCamouflage}.
In the field of computer vision, research on camouflaged objects is often associated with salient object detection (SOD), which mainly deals with those salient and easily observed objects in the scene.
In general, saliency models are designed for the general observation paradigm (\textit{i.e.}, finding visually prominent objects).
They are not suitable for the specific observation (\textit{i.e.}, finding concealed objects).
Therefore, it is necessary to establish models based on the essential requirements and specific data of the task to learn the special knowledge.

\begin{figure*}[t]
	\centering
	\includegraphics[width=1\linewidth]{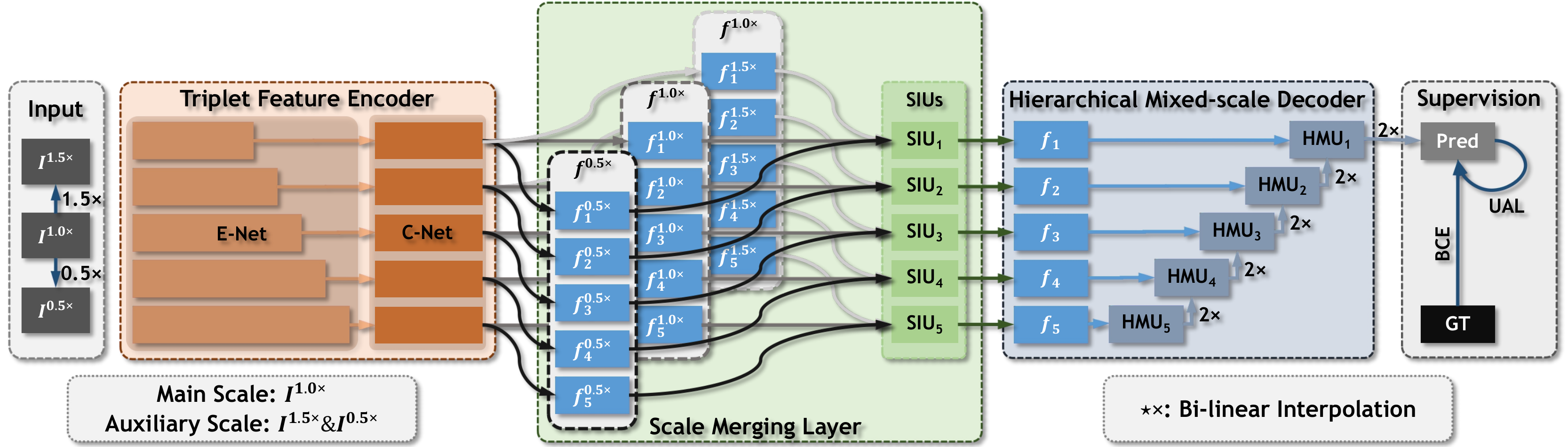}
	\caption{
		Overall framework.
		The shared triplet feature encoder is used to extract multi-level features corresponding to different input ``zoom'' scales, which is composed of E-Net and C-Net for extracting and compressing features, respectively.
		At different levels of the scale merging layer, SIUs are adopted to screen and aggregate the critical cues from different scales.
		Then the fused features are gradually integrated through the top-down up-sampling path in the hierarchical mixed-scale decoder.
		HMUs further enhance the feature discrimination by constructing a multi-path structure inside the features.
		Finally, a probability map of the camouflaged object corresponding to the input image can be obtained.
		In the training stage, the binary cross entropy and the proposed UAL are used as the loss function.
	}
	\label{fig:net}
\end{figure*}

\noindent{\textbf{Camouflaged Object Detection (COD).}}
Different from the traditional SOD task, the COD pays more attention to the undetectable objects (mainly because of too small size, occlusion, concealment or self-disguise).
Due to the differences in the attributes of the objects of interest, the goals of the two tasks are different.
The difficulty and complexity of the COD far exceed the SOD due to the high similarity between the object and the environment.
Some valuable attempts have been made in recent years.
Recent works~\cite{CAMO,COD-MGL,SLSR} construct the multi-task learning framework in the prediction process of camouflaged objects and introduce some auxiliary tasks like classification and edge detection.
Some uncertainty-aware methods~\cite{UJSC,COD-UGTR} are proposed to model and cope with the uncertainty in data annotation or COD data itself.
In the other two methods~\cite{COD-PFNet,COD-C2FNet}, contextual feature learning also plays an important role.
There are also a number of bio-inspired methods, such as~\cite{COD10K,COD-MirrorNet}.
They capture camouflaged objects by imitating the behavior process of hunters or changing the viewpoint of the scene.
Although our method can also be attributed to the last category, ours is different from the above methods.
Our method simulates the behavior of humans to understand complex images by zooming in and out strategy.
The proposed method explores the scale-specific and imperceptible semantic features under the mixed scales for accurate predictions, with the supervision of BCE and our proposed uncertainty-aware loss.
Accordingly, our method achieves a more comprehensive understanding of the scene, and accurately and robustly segments the camouflaged objects from the complex background, which even can be transferred to the SOD task effectively and smoothly.

\noindent{\textbf{Scale Space Integration.}}
The scale-space theory aims to promote an optimal understanding of image structure, which is an extremely effective and theoretically sound framework for addressing naturally occurring scale variations.
Its ideas have been widely used in computer vision, including the image pyramid~\cite{ImagePyramid} and the feature pyramid~\cite{FPN}.
Due to the structural and semantic differences at different scales, the corresponding features play different roles.
However, the commonly-used inverted pyramid-like feature extraction structures~\cite{SPP,FasterRCNN,MSAPS} often cause the feature representation to lose too much texture and appearance details, which are unfavorable for dense prediction tasks~\cite{FCN,Unet} that emphasize the integrity of regions and edges.
Thus, some recent CNN-based COD methods~\cite{COD10K,COD-C2FNet,COD-PFNet,COD-MGL} and SOD methods~\cite{PoolNet,MINet,GateNet,DMRA-TIP,HDFNet,DANet,SSLSOD} explore the combination strategy of inter-layer features to enhance the feature representation.
These bring some positive gains for accurate localization and segmentation of objects.
However, for the COD task, the existing approaches overlook the performance bottleneck caused by the ambiguity of the structural information of the data itself that makes it difficult to be fully perceived at a single scale.
Different from them, we mimic the zoom strategy to synchronously consider differentiated relationships between object and background at multiple scales, thereby fully perceiving the camouflaged objects and confusing scenes.
Besides, we also further explore the fine-grained feature scale space between channels.

\section{Proposed Method}

In this section, we first elaborate on the overall architecture of the proposed ZoomNet, and then present the details of each module and the uncertainty-aware loss.

\subsection{Overall Architecture}\label{sec:architecture}

The overall architecture of the proposed ZoomNet is illustrated in Fig.~\ref{fig:net}.
Inspired by the zoom strategy from human beings when observing confusing scenes, we argue that different zoom scales often contain their specific information.
Aggregating the differentiated information on different scales will benefit exploring the inconspicuous yet valuable clues from confusing scenarios, thus facilitating COD.
To implement it, intuitively, we resort to the image pyramid.
Specifically, we customize an image pyramid based on the single scale input to identify the camouflaged objects.
The scales are divided into a main scale (\textit{i.e.} the input scale) and two auxiliary scales.
The latter is obtained by re-scaling the former to imitate the operation of zooming in and out.
We utilize the shared triplet feature encoder to extract features on different scales and feed them to the scale merging layer.
To integrate these features that contain rich scale-specific information, we design a series of scale integration units (SIUs) based on the attention-aware filtering mechanism.
Thus, these auxiliary scales are integrated into the main scale, \textit{i.e.}, information aggregation of ``zoom in and out'' operation.
This will largely enhance the model to distill critical and informative semantic cues for capturing difficult-to-detect camouflaged objects.
After that, we construct hierarchical mixed-scale units (HMUs) to gradually integrate multi-level features in a top-down manner to enhance the mixed-scale feature representation.
It further increases the receptive field range and diversifies feature representation within the module.
The captured fine-grained and mixed-scale clues promote the model to accurately segment the camouflaged objects in the chaotic scenes.
Besides, to overcome the uncertainty in the prediction caused by the inherent complexity of the data, we design an uncertainty-aware loss (UAL) to assist the BCE loss, enabling the model to distinguish these uncertain regions and produce an accurate and reliable prediction.

\subsection{Triplet Feature Encoder}\label{sec:encoder}

We start by extracting deep features through a shared triplet feature encoder for the group-wise inputs, which consists of the feature extraction and the channel compression networks, \textit{i.e.} E-Net and C-Net.
For the trade-off between efficiency and effectiveness, the main scale and the two auxiliary scales are empirically set to $1.0\times$, $1.5\times$ and $0.5\times$.
E-Net is constituted by the commonly-used ResNet-50~\cite{Resnet} that is removed the structure after ``layer4''.
C-Net is cascaded to further optimize computation and obtain a more compact feature.
For more details about it, please see the \textit{supp}.
Thus, three sets of 64-channel feature maps corresponding to three input scales are produced, \textit{i.e.}, $\{f^{k}_{i}\}_{i=1}^{5}, \, k \in \{0.5, 1.0, 1.5\}$.
Next, these features are fed successively to the scale merging layer and the hierarchical mixed-scale decoder for subsequent processing.

\subsection{Scale Merging Layer}\label{sec:trans}

\begin{figure}[t]
	\centering
	\includegraphics[width=\linewidth]{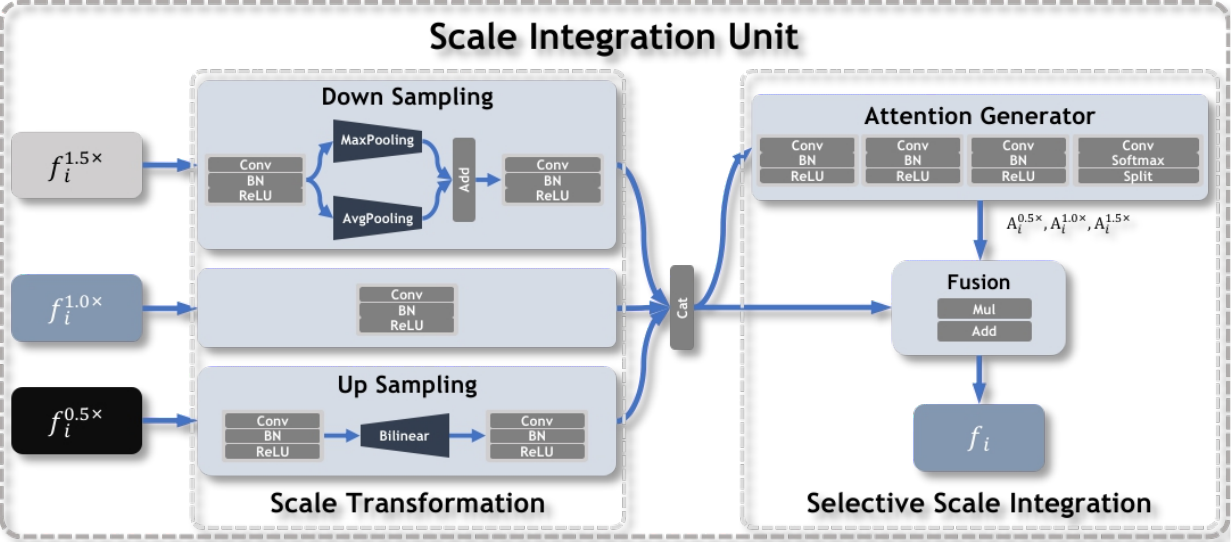}
	\caption{Illustration of the scale integration unit (SIU).}
	\label{fig:trans}
\end{figure}

We design an attention-based SIU to screen (weight) and combine scale-specific information, as shown in Fig.~\ref{fig:trans}.
Several such units make up the scale merging layer.
Through filtering and aggregation, the expression of different scales is self-adaptively highlighted.
Before scale integration, the features $f^{1.5}_{i}$ and $f^{0.5}_{i}$ are first resized to be consistent resolution with the main scale feature $f^{1.0}_{i}$.
Specifically, for $f^{1.5}_{i}$, we use a hybrid structure of ``max-pooling + average-pooling'' to down-sample it, which helps to preserve the effective and diverse responses for camouflaged objects in high-resolution features.
For $f^{0.5}_{i}$, we directly up-sample it by the bi-linear interpolation.
Then, these features are fed into the ``attention generator'', and a three-channel feature map is calculated through a series of convolutional layers.
After a softmax activation layer, the attention map $A^k$ ($k \in \{0.5, 1.0, 1.5\}$) corresponding to each scale can be obtained and used as respective weights for the final integration.
The process is formulated as:
\begin{equation} \label{equ:fusion}
	\begin{split}
		A_{i} & = \text{softmax}(\Psi( \left[\mathcal{U}(f^{0.5}_{i}), f^{1.0}_{i}, \mathcal{D}(f^{1.5}_{i}) \right], \phi)), \\
		f_{i} & = A^{0.5}_{i} \cdot \mathcal{U}(f^{0.5}_{i}) + A^{1.0}_{i} \cdot f^{1.0}_{i} + A^{1.5}_{i} \cdot \mathcal{D}(f^{1.5}_{i}),
	\end{split}
\end{equation}
where $\Psi(\star, \phi)$ indicates the stacked ``Conv-BN-ReLU'' layers in the attention generator, and $\phi$ means the parameters of these layers.
$\left[ \star \right]$ represents the concatenation operation.
$\mathcal{D}$ and $\mathcal{U}$ refer to the hybrid pooling and bi-linear interpolation operations mentioned above, respectively.
Note that some operations before and after the sampling operation are not shown in Equ.~\ref{equ:fusion} for simplicity but can be seen in Fig.~\ref{fig:trans}.
These designs aim to selectively aggregate the scale-specific information to explore subtle but critical semantic cues at different scales, boosting the feature representation.

\subsection{Hierarchical Mixed-scale Decoder}\label{sec:decoder}

\begin{figure}[t]
	\centering
	\includegraphics[width=1 \linewidth]{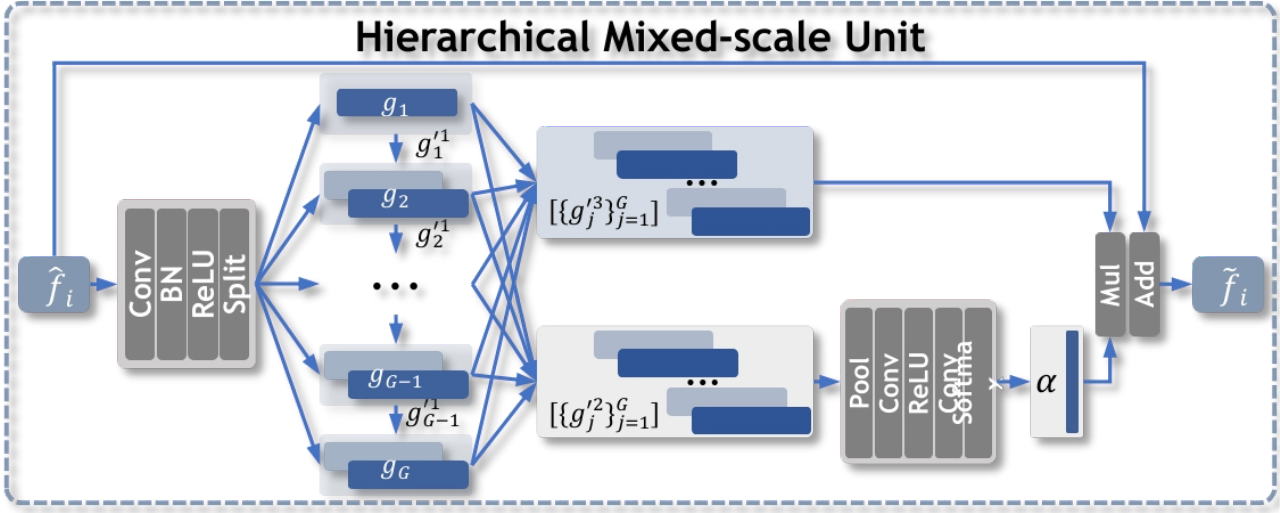}
	\caption{
		Hierarchical mixed-scale unit (HMU).
		We adopt group-wise interaction and channel-wise modulation to explore the discriminative and valuable semantics from different channels.
		Note that each group of features is executed sequentially from top to bottom.
		The latter one integrates part of the features of the previous one before the feature transformation.
	}
	\label{fig:decoder}
\end{figure}

\begin{table*}[t]
	\centering
	\caption{
		Comparisons of different methods on COD datasets.
		The best three results are highlighted in {\color{reda} \textbf{red}}, {\color{mygreen} \textbf{green}} and {\color{myblue} \textbf{blue}}.
		``\none'': Not available;
		$^{\star}$: Using more datasets.
	}
	\resizebox{\linewidth}{!}{%
		\input{tables/cod_sota.tex}
	}
	\label{tab:totalcod}
\end{table*}

After SIUs, the auxiliary-scale information is integrated into the main-scale branch.
Similar to the multi-scale case, different channels also contain differentiated semantics.
Thus, it is necessary to excavate valuable clues contained in different channels.
To this end, we design HMUs to conduct information interaction and feature refinement between channels, which strengthen features from coarse-grained group-wise iteration to fine-grained channel-wise modulation in the decoder, as depicted in Fig.~\ref{fig:decoder}.
The input $\hat{f}_{i}$ of the HMU$_i$ contains the multi-scale fused feature $f_{i}$ from the SIU$_i$ and the feature $\tilde{f}_{i+1}$ from the HMU$_{i+1}$:
\begin{equation}
	\begin{split}
		\hat{f}_{i} = f_{i} + \mathcal{U}(\tilde{f}_{i+1}).
	\end{split}
\end{equation}
\noindent \textbf{Group-wise Iteration.}
We adopt $1 \times 1$ convolution to extend the channel number of feature map $\hat{f}_{i}$.
The features are then divided into $G$ groups $\{g_{j}\}_{j=1}^{G}$ along the channel dimension.
Feature interaction between groups is carried out in an iterative manner.
Specifically, the first group $\{g_{1}\}$ is split into three feature sets $\{{g'}_{1}^{k}\}_{k=1}^{3}$ after a convolution block.
Among them, the ${g'}_{1}^{1}$ is adopted for information exchange with the next group, and the other two are used for channel-wise modulation.
In the $j^{th}$ ($1<j<G$) group, the feature $g_{j}$ is concatenated with the feature ${g'}_{j-1}^{1}$ from the previous group along the channel, followed by a convolution block and a split operation, which similarly divides this feature group into three feature sets.
It is noted that the output of the group $G$ with the similar input form to the previous groups only contains ${g'}_{G}^{2}$ and ${g'}_{G}^{3}$. % NOTE 因为第G组仅仅划分成两组输出，和其他组不同，所以这里应该强调一下
Such an iterative mixing strategy strives to learn the critical clues from different channels and obtain a powerful feature representation.
From another perspective, the iterative structure in HMU can be equivalent to a kernel pyramid structure.  % TODO 这里是之前一个reviewer问的问题，问为什么不使用核金字塔。所以说这里也没有具体的参考文献。

\noindent \textbf{Channel-wise Modulation.}
The features $[\{{g'}_{j}^{2}\}_{j=1}^{G}]$ are concatenated and converted into the feature modulation vector $\mathbf{\alpha}$ by a small convolutional network, which is employed to weight another concatenated feature $[\{{g'}_{j}^{3}\}_{j=1}^{G}]$.
The weighted feature is then processed by a convolutional layer, which is defined as:
\begin{equation}
	\begin{split}
		\tilde{f}_{i} = \mathcal{A}(\hat{f}_{i} + \mathcal{N}(\mathcal{T}(\alpha \cdot [\{{g'}_{j}^{3}\}_{j=1}^{G}]))),
	\end{split}
\end{equation}
where $\mathcal{A}$, $\mathcal{N}$ and $\mathcal{T}$ represent the activation layer, the normalization layer and the convolutional layer, respectively.

Based on five cascaded HMUs and several stacked convolutional layers, a single-channel logits map is obtained. The final confidence map $\mathbf{P}$ that highlights the camouflaged objects is then generated by a sigmoid function.

\subsection{Loss Functions}\label{sec:loss}

\begin{figure*}[t]
	\centering
	\includegraphics[width=1 \linewidth]{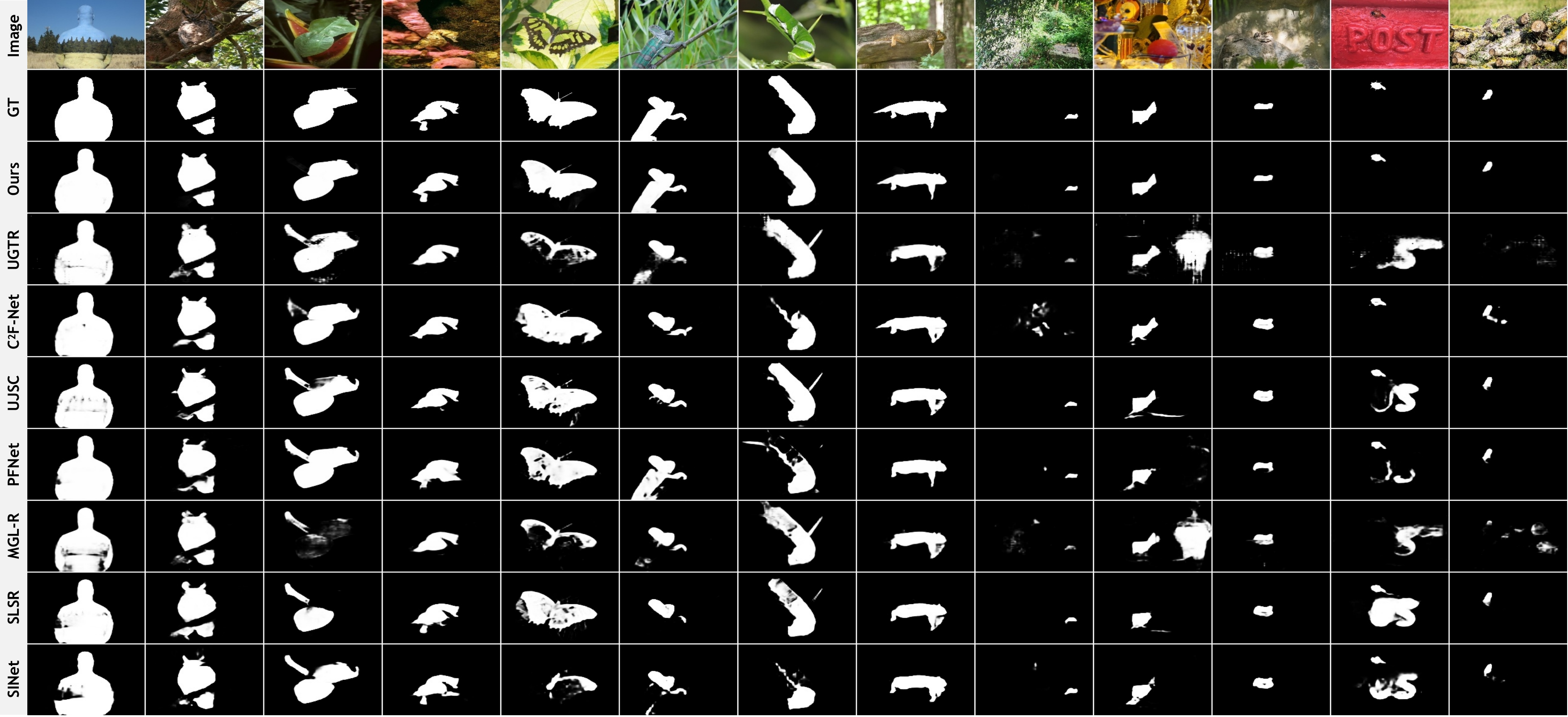}
	\caption{
		Visual comparisons of some recent COD methods and ours on different types of samples.
		Please zoom in for more details.
	}
	\label{fig:visualcomp}
\end{figure*}

The binary cross entropy loss (BCEL) is widely used in various image segmentation tasks and its mathematical form is $l_{BCEL}^{i,j} = - \mathbf{g}_{i,j} \log \mathbf{p}_{i,j} - (1 - \mathbf{g}_{i,j}) \log (1 - \mathbf{p}_{i,j})$, where $\mathbf{g}_{i,j} \in \{0, 1\}$ and $\mathbf{p}_{i,j} \in [0, 1]$ denote the ground truth and the predicted value at position $(i,j)$, respectively.
As shown in Fig.~\ref{fig:visualation}, due to the complexity of the COD data, if trained only under the BCEL, the model produces serious ambiguity and uncertainty in the prediction and fails to accurately capture objects, of which both will reduce the reliability of COD.
To force the model to enhance ``confidence'' in decision-making and increase the penalty for fuzzy prediction, we design a strong constraint as the auxiliary of the BCEL, \textit{i.e.}, the uncertainty-aware loss (UAL).

In the final probability map of the camouflaged object, the pixel value range is $[0, 1]$, where $0$ means the pixel belongs to the background, and $1$ means it belongs to the camouflaged object.
Therefore, the closer the predicted value is to $0.5$, the more uncertain the determination about the property of the pixel is.
To optimize it, a direct way is to use the ambiguity as the supplementary loss for these difficult samples.
To this end, we first need to define the ambiguity measure of the pixel $x$, which maximizes at $x=0.5$ and minimizes at $x=0$ or $x=1$.
And as a loss, the function should be smooth and continuous with only a finite number of non-differentiable points.
For brevity, we empirically consider two forms, $\Phi_{pow}^{\alpha}(x)=1-|2x-1|^{\alpha}$ based on the power function and $\Phi_{exp}^{\alpha}(x)=e^{-(\alpha (x-0.5))^{2}}$ based on the exponential function.
Besides, inspired by the form of the weighted BCE loss, we also try to use $\omega = 1+\Phi_{pow}^{2}(x)$ as the weight of BCE loss to increase the loss of hard pixels.
After massive experiments (Sec.~\ref{sec:loss_form}), the proposed UAL is formulated as $l_{UAL}^{i,j} = 1-|2\mathbf{p}_{i,j}-1|^{2}$.
Finally, the total loss function can be written as:
\begin{equation}
	\begin{split}
		L = L_{BCEL} + \lambda \times L_{UAL},
	\end{split}
\end{equation}
where $\lambda$ is the balance coefficient and we design three adjustment strategies of $\lambda$, \textit{i.e.}, a fixed constant value, an increasing linear strategy, and an increasing cosine strategy in Sec.~\ref{sec:ablation_coef}.
The different forms and corresponding results are listed in the supp.
From the results, we find that the increasing strategies, especially ``cosine'', do achieve better performance.
So, the cosine strategy is used by default.

\section{Experiments}

\subsection{Experiment Setup}

\noindent \textbf{Datasets.}
We use four COD datasets, CAMO~\cite{CAMO}, CHAMELEON~\cite{CHAMELEON}, COD10K~\cite{COD10K} and NC4K~\cite{SLSR}.
CAMO consists of 1,250 camouflaged and 1,250 non-camouflaged images.
CHAMELEON contains 76 hand-annotated images.
COD10K includes 5,066 camouflaged, 3,000 background and 1,934 non-camouflaged images.
NC4K is another large-scale COD testing dataset including 4,121 images from the Internet.
Following the data partition of~\cite{COD10K,SLSR,COD-MGL,COD-PFNet}, we use all images with camouflaged objects in the experiments, in which 3,040 images from COD10K and 1,000 images from CAMO are used for training, and the rest ones for testing.
\textit{Besides, we also show the performance on five SOD datasets in the supp.}

\noindent \textbf{Evaluation Criteria.}
For COD and SOD, we use seven common metrics for evaluation based on~\cite{PySODMetrics, PySODEvalToolkit}, including
S-measure~\cite{Smeasure} (S$_{m}$),
weighted F-measure~\cite{wFmeasure} (F$_{\beta}^{\omega}$),
mean absolute error (MAE),
F-measure~\cite{Fmeasure} (F$_{\beta}$),
E-measure~\cite{Emeasure} (E$_{m}$),
precision-recall (PR) curve
and F$_{\beta}$-threshold curve (F$_{\beta}$ curve).
The curves can be found in the \textit{supp}.

\noindent \textbf{Implementation Details.}
The proposed ZoomNet is implemented with PyTorch.
As the settings in recent methods~\cite{COD10K,SLSR,COD-MGL,COD-PFNet}, the encoder is initialized with the parameters of ResNet-50 pretrained on ImageNet, and the remaining parts are randomly initialized.
SGD with momentum 0.9 and weight decay 0.0005 is chosen as the optimizer.
The learning rate is initialized to 0.05 and follows a linear warm-up and linear decay strategy.
The entire model is trained for 40 epochs with a batch size of 8 in an end-to-end manner on an NVIDIA 2080Ti GPU.
During training and inference, the main scale is $384 \times 384$.
Random flipping and rotating are employed to augment the training data.

\subsection{Comparisons with State-of-the-arts}

COD is an emerging field, so we introduce some methods for salient object detection and medical image segmentation for comparison.
The results of all these methods come from existing public data or are generated by models that are retrained based on the code released by the authors.

\noindent \textbf{Quantitative Evaluation.}
Tab.~\ref{tab:totalcod} shows the detailed comparison results.
It can be seen that the proposed model consistently and significantly surpasses recent methods on all datasets without relying on any post-processing tricks.
Compared with the recent best COD method UJSC, although it introduces extra SOD data for training and has suppressed other existing methods, our method still shows the obvious performance improvement on these datasets.
Especially, our approach has more advantages on the metrics F$_{\beta}^{\omega}$, MAE, and F$_{\beta}$.
On four datasets, the proposed method averagely outperforms the second-best method C$^2$F-Net by $19.3\%$ in terms of MAE and the average gains in terms of F$_{\beta}^{\omega}$ and F$_{\beta}$ are $4\%$.
Besides, PR and F$_{\beta}$ curves shown in the \textit{supp.} also demonstrate the effectiveness of the proposed method.
The flatness of the F$_{\beta}$ curve reflects the consistency and uniformity of the prediction.
Our curves are almost horizontal, which can be attributed to the effect of the proposed UAL.
It drives the predictions to be more polarized and reduces the ambiguity.

\begin{table}[t]
	\centering
	\caption{
		Ablation study on the COD10K-Test.
		SIU: Scale integration unit;
		HMU: Hierarchical mixed-scale unit with $g$ groups;
		UAL: Uncertainty-aware loss;
		\ding{176}: The simple extension of baseline \ding{172} with the similar number of parameters and FLOPs to \ding{175}.
	}
	\resizebox{\linewidth}{!}{%
		\input{tables/ablation_component.tex}
	}
	\label{tab:ablationstudy}
\end{table}

\noindent \textbf{Qualitative Evaluation.}
Visual comparisons of different methods on several typical samples are shown in Fig.~\ref{fig:visualcomp}.
They present the complexity in different aspects, such as
big objects (Col. 1),
middle objects (Col. 2-8),
small objects (Col. 9-13),
occlusions (Col. 2 and 10),
background interference (Col. 10-13),
and indefinable boundaries (Col. 1, 2, 6-13).
These results intuitively show the superior performance of the proposed method.
In addition, it can be noticed that our predictions have clearer and more complete object regions and sharper contours.

\subsection{Ablation Studies}\label{ablation}

In this section, we perform comprehensive ablation analyses on different components.
Because COD10K is the most widely-used large-scale COD dataset, and contains various objects and scenes, all subsequent ablation experiments are carried out on it.

\begin{figure}[t]
	\centering
	\includegraphics[width=1 \linewidth]{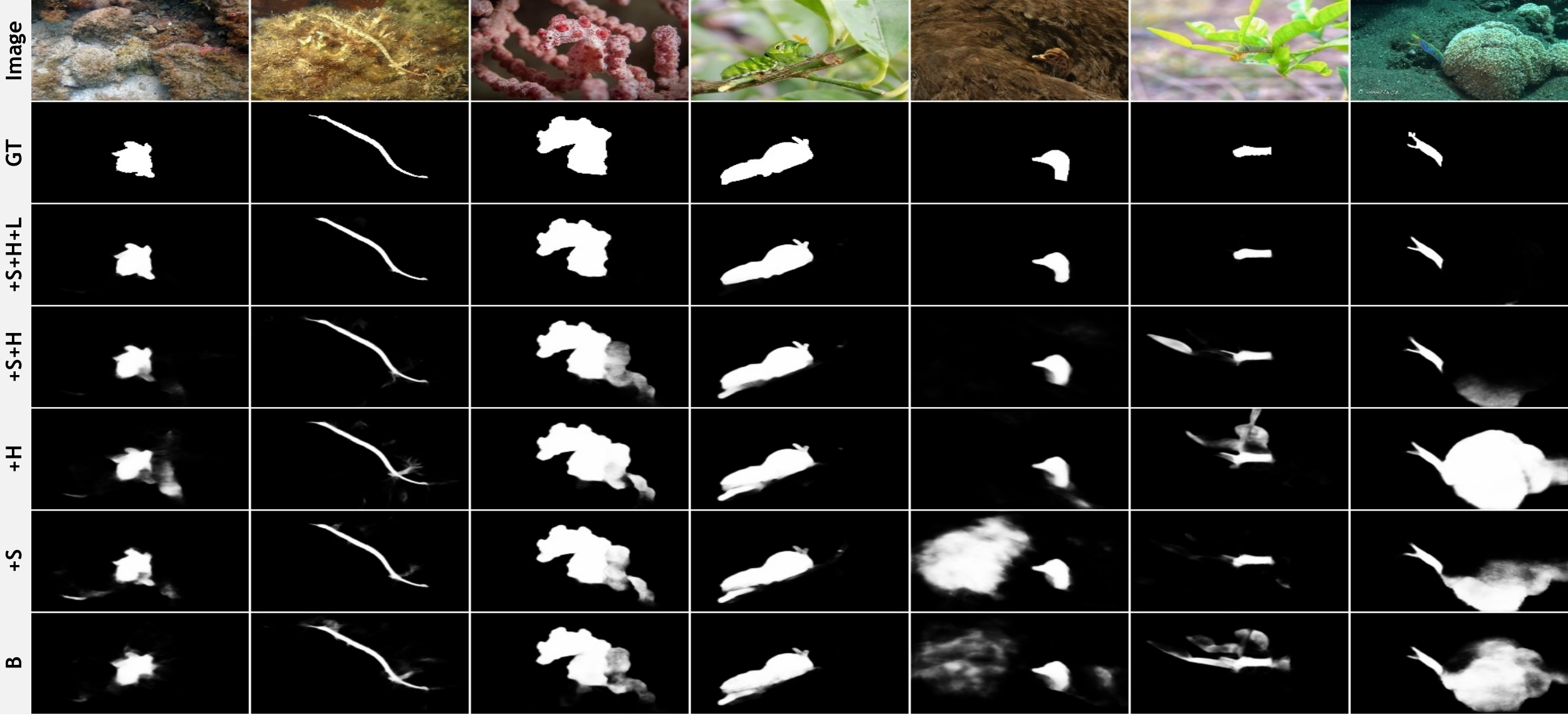}
	\caption{
		Visual comparisons for showing the effects of the proposed components.
		B: Baseline;
		+S: +SIUs;
		+H: +HMUs;
		+S+H: +SIUs+HMUs;
		+S+H+L: +SIUs+HMUs+UAL.
	}
	\label{fig:visualation}
\end{figure}

\begin{table}[t]
	\centering
	\caption{
		Comparisons of mixed and single scale input schemes on CAMO and COD10K.
		All models are based on \ding{172} in Tab.~\ref{tab:ablationstudy}.
	}
	\resizebox{\linewidth}{!}{%
		\input{tables/msi.tex}
	}
	\label{tab:msi_scheme}
\end{table}

\noindent \textbf{Effectiveness of SIUs and HMUs.}\label{sec:ablation_siuhmu}
In the proposed model, both the SIU and the HMU are very important structures.
We install them one by one on the baseline model to evaluate their performance.
The results are shown in Tab.~\ref{tab:ablationstudy}.
Our baseline \ding{172} and other models \ding{173} and \ding{176} only use the inputs of the main scale.
As can be seen, our baseline shows a good performance, probably due to the proper training setup and the more reasonable network architecture detailed in the \textit{supp}.
From \ding{172}-\ding{175}, it can be seen that the two proposed modules make a significant contribution to the performance when compared to the baseline.
Besides, the results in Fig.~\ref{fig:visualation} show that the two modules can benefit each other and reduce their errors (\textit{e.g.}, Col. 1, 2, 5 and 7) to locate and distinguish objects more accurately.
These components effectively help the model to excavate and distill the critical and valuable semantics and improve the capability of distinguishing hard objects. % more accurately understand the context information
Under the cooperation between the proposed modules and loss functions, ZoomNet can completely capture the camouflaged objects of different scales and generate the predictions with higher contrast and consistency.
In addition, in Tab.~\ref{tab:ablationstudy}, the model \ding{176} is a simple extension of the baseline model \ding{172} using some standard convolutional blocks, to make the similar number of parameters and FLOPs with \ding{175}.
The model \ding{175} still achieves better performance, which reflects the effectiveness of the proposed modules and the rationality of the design.

\noindent \textbf{Number of Groups in HMUs.}\label{sec:num_groups}
In Tab.~\ref{tab:ablationstudy}, we also show the effects of different group numbers in the proposed HMU.
It can be seen from the results that the best performance appears when the number of groups is equal to 6.
Also, it achieves a good balance between performance and efficiency.
So, in other experiments, we set the number of groups in each HMU to 6.

\noindent \textbf{Mixed-scale Input Scheme.}\label{sec:msi}
Our model is designed to mimic the behavior of "Zoom In\&Out".
The feature expression is enriched by combining the scale-specific information from different scales.
In Fig.~\ref{fig:top_img}, the intermediate features and attention maps show that our mixed-scale scheme plays a positive and important role in locating the camouflaged object.
Considering that the objects in COD10K are mainly small objects~\cite{COD10K}, which may limit the role of 0.5$\times$ input to some extent, we list average results on COD10K and CAMO in Tab.~\ref{tab:msi_scheme}.
The proposed scheme performs better than the single-scale one and simply mixed one.
This verifies the rationality of such a design for the COD task.

\noindent \textbf{Options of Setting $\lambda$.}\label{sec:ablation_coef}
We compare three strategies and the results are listed in the supp., in which the increasing cosine strategy achieves the best performance.
This may be due to the advantage of its smooth change process.
This smooth intensity warm-up strategy of UAL motivates the model to take advantage of UAL in improving the learning process and to mitigate the possible negative interference of UAL on BCEL due to the lower accuracy of the model during the early stage of training.

\begin{table}[t]
	\centering
	\caption{
		Different forms of the proposed UAL.
		Form 0 is our model without UAL.
		``\none'': Unable to converge.
		Form 1.5 is used by default because of its balanced performance.
		Their curves are shown in the \textit{supp}.
	}
	\resizebox{\linewidth}{!}{%
		\input{tables/loss_form.tex}
	}
	\label{tab:loss_form}
\end{table}

\noindent \textbf{Forms of UAL.}\label{sec:loss_form}
Different forms of UAL are listed in Tab.~\ref{tab:loss_form} and the corresponding curves are illustrated in the \textit{supp}.
As can be seen, Form 1.5 has a more balanced performance.
Also, it is worth noting that, when approaching 0 or 1, the form which can maintain a larger gradient will obtain better performance in terms of F$^{\omega}_{\beta}$, MAE and F$_{\beta}$.
This may provide some reference for designing a better loss.

\noindent \textbf{Effectiveness of UAL.}\label{sec:ual}
The results of Fig.~\ref{fig:visualation} intuitively show that the UAL greatly reduces the ambiguity caused by the interference from the background.
Besides, we visualize the histogram maps of all results on CHAMELEON and the intermediate features from different stages in the decoder in the \textit{supp}.
In the stacked histogram map ``w/o UAL'', a large number of pixels appear in the middle area, which corresponds to more visually blurred/uncertain predictions.
Besides, in the corresponding feature visualization, especially in the region inside the red box, there is clear background interference due to the complex scenarios and blurred edges, which are extremely prone to yield false positive predictions.
However, when UAL is introduced, it can be seen that the middle interval of ``w UAL'' is flatter than the one of ``w/o UAL'', that is, most pixel values approach two extremes.
And the feature maps become more discriminative and present a more compact and complete response in the regions of camouflaged objects.

\subsection{Discussion on SOD and COD}
It can be seen from the experiments in the \textit{supp.} that our method not only performs well on COD, but also shows outstanding performance on SOD.
Considering the difference between these two tasks, we may wonder \textit{why our method consistently performs well on such two seemingly different tasks}.
We attribute this to the generality and rationality of the designed structure.
In fact, SOD and COD have a clear commonality, \textit{i.e.}, the accurate segmentation has to depend on multi-scale and category-free discriminative features.
By integrating rich scale-specific features, our model can extract critical and informative cues from scenes and objects, which helps precise localization and smooth segmentation of objects.
In addition, the proposed UAL can mitigate the ambiguity of predictions caused by the inherent complexity of scenes.
Although the objects in SOD are salient, it can also benefit from UAL due to the vagueness introduced by the CNN model itself in the detailed information recovery process.
All of these components are built on the common demand of the two tasks, which provides a solid foundation for the performance.

\section{Conclusion}

In this paper, we propose the ZoomNet by imitating the behavior of human beings to zoom in and out on images.
This process actually considers the differentiated expressions about the scene from different scales, which helps to improve the understanding and judgment of camouflaged objects.
We first filter and aggregate scale-specific features through the scale merging layer to enhance feature representation.
Next, in the hierarchical mixed-scale decoder, the strategies of grouping, mixing and fusion further mine the mixed-scale semantics.
Lastly, we introduce the uncertainty-aware loss to penalize the ambiguity of the prediction.
Extensive experiments verify the effectiveness of the proposed method in both the COD and SOD tasks with superior performance to existing state-of-the-art methods.

\noindent\textbf{Acknowledgements}
This work was supported by the National Natural Science Foundation of China \#61876202 and \#61829102, the Liaoning Natural Science Foundation \#2021-KF-12-10, and the Fundamental Research Funds for the Central Universities \#DUT20ZD212.

\input{section/appendix.tex}

%%%%%%%%% REFERENCES
{\small
	\bibliographystyle{ieee_fullname}
	\bibliography{egbib}
}

\end{document}

%% file: tables/cod_sota.tex
\begin{tabular}{l|ccccc|ccccc|ccccc|ccccc}
 \toprule[2pt]
 \rowcolor{tabtitle}
                                 & \multicolumn{5}{c|}{\textbf{CAMO}} & \multicolumn{5}{c|}{\textbf{CHAMELEON}} & \multicolumn{5}{c}{\textbf{COD10K}} & \multicolumn{5}{c}{\textbf{NC4K}}                                                                                                                                                                                                                                                                                                                                                                                                                                                                                                                                                                                 \\ \rowcolor{tabtitle}
 \multirow{-2}{*}{Model}         & S$_{m}$ $\uparrow$                 & F$^{\omega}_{\beta}$ $\uparrow$         & MAE $\downarrow$                    & F$_{\beta}$ $\uparrow$            & E$_{m}$ $\uparrow$               & S$_{m}$ $\uparrow$               & F$^{\omega}_{\beta}$ $\uparrow$  & MAE $\downarrow$                 & F$_{\beta}$ $\uparrow$           & E$_{m}$ $\uparrow$               & S$_{m}$ $\uparrow$               & F$^{\omega}_{\beta}$ $\uparrow$  & MAE $\downarrow$                 & F$_{\beta}$ $\uparrow$           & E$_{m}$ $\uparrow$               & S$_{m}$ $\uparrow$               & F$^{\omega}_{\beta}$ $\uparrow$  & MAE $\downarrow$                 & F$_{\beta}$ $\uparrow$           & E$_{m}$ $\uparrow$               \\ \midrule[1pt]
 \multicolumn{21}{c}{\textbf{Salient Object Detection / Medical Image Segmentation}}                                                                                                                                                                                                                                                                                                                                                                                                                                                                                                                                                                                                                                                                                      \\ \midrule[1pt]
 NLDF~\cite{NLDF}                & 0.665                              & 0.495                                   & 0.123                               & 0.564                             & 0.790                            & 0.798                            & 0.652                            & 0.063                            & 0.714                            & 0.893                            & 0.701                            & 0.473                            & 0.059                            & 0.539                            & 0.819                            & 0.738                            & 0.586                            & 0.083                            & 0.656                            & 0.846                            \\
 PiCANet~\cite{PiCANet}          & 0.701                              & 0.510                                   & 0.125                               & 0.573                             & 0.799                            & 0.765                            & 0.552                            & 0.085                            & 0.618                            & 0.846                            & 0.696                            & 0.415                            & 0.081                            & 0.489                            & 0.788                            & 0.758                            & 0.570                            & 0.088                            & 0.640                            & 0.835                            \\
 BASNet~\cite{BASNet}            & 0.615                              & 0.434                                   & 0.124                               & 0.503                             & 0.727                            & 0.847                            & 0.771                            & 0.044                            & 0.795                            & 0.894                            & 0.661                            & 0.432                            & 0.071                            & 0.486                            & 0.749                            & 0.695                            & 0.546                            & 0.095                            & 0.610                            & 0.785                            \\
 CPD~\cite{CPD}                  & 0.716                              & 0.556                                   & 0.113                               & 0.618                             & 0.796                            & 0.857                            & 0.731                            & 0.048                            & 0.771                            & 0.923                            & 0.750                            & 0.531                            & 0.053                            & 0.595                            & 0.853                            & 0.787                            & 0.645                            & 0.072                            & 0.705                            & 0.866                            \\
 PoolNet~\cite{PoolNet}          & 0.730                              & 0.575                                   & 0.105                               & 0.643                             & 0.819                            & 0.845                            & 0.690                            & 0.054                            & 0.749                            & 0.933                            & 0.740                            & 0.506                            & 0.056                            & 0.575                            & 0.844                            & 0.785                            & 0.635                            & 0.073                            & 0.699                            & 0.865                            \\
 EGNet~\cite{EGNet}              & 0.732                              & 0.604                                   & 0.109                               & 0.670                             & 0.820                            & 0.797                            & 0.649                            & 0.065                            & 0.702                            & 0.884                            & 0.736                            & 0.517                            & 0.061                            & 0.582                            & 0.854                            & 0.777                            & 0.639                            & 0.075                            & 0.696                            & 0.864                            \\
 F3Net~\cite{F3Net}              & 0.711                              & 0.564                                   & 0.109                               & 0.616                             & 0.780                            & 0.848                            & 0.744                            & 0.047                            & 0.770                            & 0.917                            & 0.739                            & 0.544                            & 0.051                            & 0.593                            & 0.819                            & 0.780                            & 0.656                            & 0.070                            & 0.705                            & 0.848                            \\
 SCRN~\cite{SCRN}                & 0.779                              & 0.643                                   & 0.090                               & 0.705                             & 0.850                            & 0.876                            & 0.741                            & 0.042                            & 0.787                            & 0.939                            & 0.789                            & 0.575                            & 0.047                            & 0.651                            & 0.880                            & 0.830                            & 0.698                            & 0.059                            & 0.757                            & 0.897                            \\
 CSNet~\cite{SOD-CSNet(SOD100K)} & 0.771                              & 0.641                                   & 0.092                               & 0.705                             & 0.849                            & 0.856                            & 0.718                            & 0.047                            & 0.766                            & 0.928                            & 0.778                            & 0.569                            & 0.047                            & 0.634                            & 0.871                            & 0.750                            & 0.603                            & 0.088                            & 0.655                            & 0.793                            \\
 SSAL~\cite{SSAL}                & 0.644                              & 0.493                                   & 0.126                               & 0.579                             & 0.780                            & 0.757                            & 0.639                            & 0.071                            & 0.702                            & 0.856                            & 0.668                            & 0.454                            & 0.066                            & 0.527                            & 0.789                            & 0.699                            & 0.561                            & 0.093                            & 0.644                            & 0.812                            \\
 UCNet~\cite{UCNet-CVPR20}       & 0.739                              & 0.640                                   & 0.094                               & 0.700                             & 0.820                            & 0.880                            & 0.817                            & 0.036                            & 0.836                            & 0.941                            & 0.776                            & 0.633                            & 0.042                            & 0.681                            & 0.867                            & 0.811                            & 0.729                            & 0.055                            & 0.775                            & 0.886                            \\
 MINet~\cite{MINet}              & 0.748                              & 0.637                                   & 0.090                               & 0.691                             & 0.838                            & 0.855                            & 0.771                            & 0.036                            & 0.802                            & 0.937                            & 0.770                            & 0.608                            & 0.042                            & 0.657                            & 0.859                            & 0.812                            & 0.720                            & 0.056                            & 0.764                            & 0.887                            \\
 ITSD~\cite{ITSD}                & 0.750                              & 0.610                                   & 0.102                               & 0.663                             & 0.830                            & 0.814                            & 0.662                            & 0.057                            & 0.705                            & 0.901                            & 0.767                            & 0.557                            & 0.051                            & 0.615                            & 0.861                            & 0.811                            & 0.679                            & 0.064                            & 0.729                            & 0.883                            \\
 PraNet~\cite{PraNet}            & 0.769                              & 0.663                                   & 0.094                               & 0.710                             & 0.837                            & 0.860                            & 0.763                            & 0.044                            & 0.789                            & 0.935                            & 0.789                            & 0.629                            & 0.045                            & 0.671                            & 0.879                            & 0.822                            & 0.724                            & 0.059                            & 0.763                            & 0.888                            \\ \midrule[1pt]
 \multicolumn{21}{c}{\textbf{Camouflaged Object Detection}}                                                                                                                                                                                                                                                                                                                                                                                                                                                                                                                                                                                                                                                                                                               \\ \midrule[1pt]
 ANet\_SRM~\cite{CAMO}           & 0.682                              & 0.484                                   & 0.126                               & 0.541                             & 0.722                            & \none                            & \none                            & \none                            & \none                            & \none                            & \none                            & \none                            & \none                            & \none                            & \none                            & \none                            & \none                            & \none                            & \none                            & \none                            \\
 SINet~\cite{COD10K}             & 0.745                              & 0.644                                   & 0.092                               & 0.702                             & 0.829                            & 0.872                            & 0.806                            & 0.034                            & 0.827                            & 0.946                            & 0.776                            & 0.631                            & 0.043                            & 0.679                            & 0.874                            & 0.808                            & 0.723                            & 0.058                            & 0.769                            & 0.883                            \\
 SLSR~\cite{SLSR}                & 0.787                              & 0.696                                   & 0.080                               & 0.744                             & 0.854                            & 0.890                            & 0.822                            & {\color{mygreen} \textbf{0.030}} & 0.841                            & {\color{myblue} \textbf{0.948}}  & 0.804                            & 0.673                            & 0.037                            & 0.715                            & {\color{myblue} \textbf{0.892}}  & {\color{myblue} \textbf{0.840}}  & {\color{myblue} \textbf{0.766}}  & {\color{myblue} \textbf{0.048}}  & {\color{myblue} \textbf{0.804}}  & {\color{mygreen} \textbf{0.907}} \\
 MGL-R~\cite{COD-MGL}            & 0.775                              & 0.673                                   & 0.088                               & 0.726                             & 0.842                            & {\color{mygreen} \textbf{0.893}} & 0.812                            & {\color{myblue} \textbf{0.031}}  & 0.833                            & 0.941                            & {\color{myblue} \textbf{0.814}}  & 0.666                            & {\color{mygreen} \textbf{0.035}} & 0.710                            & 0.890                            & 0.833                            & 0.739                            & 0.053                            & 0.782                            & 0.893                            \\
 PFNet~\cite{COD-PFNet}          & 0.782                              & 0.695                                   & 0.085                               & 0.746                             & 0.855                            & 0.882                            & 0.810                            & 0.033                            & 0.828                            & 0.945                            & 0.800                            & 0.660                            & 0.040                            & 0.701                            & 0.890                            & 0.829                            & 0.745                            & 0.053                            & 0.784                            & 0.898                            \\
 UJSC$^{\star}$~\cite{UJSC}      & {\color{mygreen} \textbf{0.800}}   & {\color{mygreen} \textbf{0.728}}        & {\color{mygreen} \textbf{0.073}}    & {\color{mygreen} \textbf{0.772}}  & {\color{mygreen} \textbf{0.873}} & {\color{myblue} \textbf{0.891}}  & {\color{mygreen} \textbf{0.833}} & {\color{mygreen} \textbf{0.030}} & {\color{mygreen} \textbf{0.847}} & {\color{mygreen} \textbf{0.955}} & 0.809                            & {\color{myblue} \textbf{0.684}}  & {\color{mygreen} \textbf{0.035}} & {\color{myblue} \textbf{0.721}}  & 0.891                            & {\color{mygreen} \textbf{0.842}} & {\color{mygreen} \textbf{0.771}} & {\color{mygreen} \textbf{0.047}} & {\color{mygreen} \textbf{0.806}} & {\color{mygreen} \textbf{0.907}} \\
 MirrorNet~\cite{COD-MirrorNet}  & 0.785                              & {\color{myblue} \textbf{0.719}}         & {\color{myblue} \textbf{0.077}}     & 0.754                             & 0.850                            & \none                            & \none                            & \none                            & \none                            & \none                            & \none                            & \none                            & \none                            & \none                            & \none                            & \none                            & \none                            & \none                            & \none                            & \none                            \\
 C$^2$FNet~\cite{COD-C2FNet}        & {\color{myblue} \textbf{0.796}}    & {\color{myblue} \textbf{0.719}}         & 0.080                               & {\color{myblue} \textbf{0.762}}   & {\color{myblue} \textbf{0.864}}  & 0.888                            & {\color{myblue} \textbf{0.828}}  & 0.032                            & {\color{myblue} \textbf{0.844}}  & 0.946                            & 0.813                            & {\color{mygreen} \textbf{0.686}} & {\color{myblue} \textbf{0.036}}  & {\color{mygreen} \textbf{0.723}} & {\color{mygreen} \textbf{0.900}} & 0.838                            & 0.762                            & 0.049                            & 0.795                            & {\color{myblue} \textbf{0.904}}  \\
 UGTR~\cite{COD-UGTR}            & 0.784                              & 0.684                                   & 0.086                               & 0.735                             & 0.851                            & 0.888                            & 0.794                            & {\color{myblue} \textbf{0.031}}  & 0.819                            & 0.940                            & {\color{mygreen} \textbf{0.817}} & 0.666                            & {\color{myblue} \textbf{0.036}}  & 0.711                            & 0.890                            & 0.839                            & 0.746                            & 0.052                            & 0.787                            & 0.899                            \\ \rowcolor{ours}
 Ours                            & {\color{reda} \textbf{0.820}}      & {\color{reda} \textbf{0.752}}           & {\color{reda} \textbf{0.066}}       & {\color{reda} \textbf{0.794}}     & {\color{reda} \textbf{0.892}}    & {\color{reda} \textbf{0.902}}    & {\color{reda} \textbf{0.845}}    & {\color{reda} \textbf{0.023}}    & {\color{reda} \textbf{0.864}}    & {\color{reda} \textbf{0.958}}    & {\color{reda} \textbf{0.838}}    & {\color{reda} \textbf{0.729}}    & {\color{reda} \textbf{0.029}}    & {\color{reda} \textbf{0.766}}    & {\color{reda} \textbf{0.911}}    & {\color{reda} \textbf{0.853}}    & {\color{reda} \textbf{0.784}}    & {\color{reda} \textbf{0.043}}    & {\color{reda} \textbf{0.818}}    & {\color{reda} \textbf{0.912}}    \\ \bottomrule[2pt]
\end{tabular}%

%% file: tables/ablation_component.tex
\begin{tabular}{c|cc|ccc|ccccc}
 \toprule[2pt]
 \rowcolor{tabtitle}
 Model      & GFLOPs  & Params. (M) & HMU   & SIU  & UAL  & S$_{m}$ $\uparrow$ & F$^{\omega}_{\beta}$ $\uparrow$ & MAE $\downarrow$ & F$_{\beta}$ $\uparrow$ & E$_{m}$ $\uparrow$ \\ \midrule[1pt]
 \ding{172} & 41.885  & 30.453      &       &      &      & 0.812              & 0.637                           & 0.039            & 0.692                  & 0.898              \\
 \ding{173} & 78.560  & 30.696      & $g=6$ &      &      & 0.820              & 0.654                           & 0.037            & 0.706                  & 0.897              \\
 \ding{174} & 166.821 & 30.453      &       & \yes &      & 0.837              & 0.682                           & 0.034            & 0.731                  & 0.912              \\
 \ding{175} & 203.496 & 28.794      & $g=6$ & \yes &      & 0.843              & 0.694                           & 0.032            & 0.743                  & 0.909              \\
 \ding{176} & 207.647 & 41.975      &       &      &      & 0.815              & 0.632                           & 0.040            & 0.689                  & 0.896              \\ \midrule[0.5pt]
 \ding{177} & 173.616 & 30.819      & $g=2$ & \yes & \yes & 0.835              & 0.721                           & 0.030            & 0.758                  & 0.902              \\
 \ding{177} & 188.556 & 31.600      & $g=4$ & \yes & \yes & 0.836              & 0.723                           & 0.029            & 0.761                  & 0.905              \\ \rowcolor{ours}
 \ding{177} & 203.496 & 32.382      & $g=6$ & \yes & \yes & 0.838              & 0.729                           & 0.029            & 0.766                  & 0.911              \\
 \ding{177} & 218.436 & 33.163      & $g=8$ & \yes & \yes & 0.836              & 0.726                           & 0.029            & 0.763                  & 0.907              \\
 \bottomrule[2pt]
\end{tabular}%

%% file: tables/msi.tex
\begin{tabular}{cc|ccccc|c}
 \toprule[2pt]
 \rowcolor{tabtitle}
 Input Scale                       & Combination Strategy & S$_{m}$ $\uparrow$ & F$^{\omega}_{\beta}$ $\uparrow$ & MAE $\downarrow$ & F$_{\beta}$ $\uparrow$ & E$_{m}$ $\uparrow$ & AVG. Relative Improvement \\ \midrule[1pt]
 $1.0\times$                       & \none                & 0.797              & 0.649                           & 0.063            & 0.704                  & 0.875              &                           \\
 $0.5\times$                       & \none                & 0.746              & 0.553                           & 0.076            & 0.616                  & 0.833              & $\downarrow$12.03\%       \\
 $0.5\times, 1.0\times$            & Addition             & 0.801              & 0.647                           & 0.062            & 0.702                  & 0.876              & $\uparrow$0.31\%          \\
 $0.5\times, 1.0\times$            & SIU                  & 0.806              & 0.658                           & 0.059            & 0.709                  & 0.879              & $\uparrow$1.86\%          \\
 $1.5\times$                       & \none                & 0.820              & 0.683                           & 0.059            & 0.737                  & 0.890              & $\uparrow$4.05\%          \\
 $0.5\times, 1.5\times$            & Addition             & 0.820              & 0.680                           & 0.058            & 0.735                  & 0.894              & $\uparrow$4.43\%          \\
 $0.5\times, 1.5\times$            & SIU                  & 0.822              & 0.687                           & 0.056            & 0.740                  & 0.893              & $\uparrow$5.33\%          \\
 $1.0\times, 1.5\times$            & Addition             & 0.819              & 0.685                           & 0.058            & 0.738                  & 0.892              & $\uparrow$4.47\%          \\
 $1.0\times, 1.5\times$            & SIU                  & 0.826              & 0.697                           & 0.056            & 0.745                  & 0.897              & $\uparrow$5.97\%          \\
 $0.5\times, 1.0\times, 1.5\times$ & Addition             & 0.821              & 0.690                           & 0.056            & 0.742                  & 0.894              & $\uparrow$5.43\%          \\
 \rowcolor{ours}
 $0.5\times, 1.0\times, 1.5\times$ & SIU                  & 0.827              & 0.700                           & 0.054            & 0.751                  & 0.898              & $\uparrow$6.91\%          \\
 \bottomrule[2pt]
\end{tabular}%

%% file: tables/loss_form.tex
% \usepackage{booktabs}
\begin{tabular}{c|c|c|ccccc}
 \toprule[2pt]
 \rowcolor{tabtitle}
 No.           & Form                                                                 & $\alpha$      & S$_{m}$ $\uparrow$            & F$^{\omega}_{\beta}$ $\uparrow$ & MAE $\downarrow$              & F$_{\beta}$ $\uparrow$        & E$_{m}$ $\uparrow$            \\ \midrule[1pt]
 0             & \no                                                                  & \no           & 0.843                         & 0.694                           & 0.032                         & 0.743                         & 0.909                         \\ \midrule[1pt]
 1.1, 1.2, 1.3 &                                                                      & 1/8, 1/4, 1/2 & \none                         & \none                           & \none                         & \none                         & \none                         \\
 1.4           &                                                                      & 1             & 0.834                         & 0.716                           & {\color{reda} \textbf{0.029}} & 0.757                         & 0.903                         \\
 1.5           &                                                                      & 2             & 0.838                         & {\color{reda} \textbf{0.729}}   & {\color{reda} \textbf{0.029}} & {\color{reda} \textbf{0.766}} & 0.911                         \\
 1.6           &                                                                      & 4             & 0.834                         & 0.727                           & {\color{reda} \textbf{0.029}} & 0.763                         & 0.903                         \\
 1.7           & \multirow{-5}{*}{$\Phi_{pow}^{\alpha}(x)=1-|2x-1|^{\alpha}$}         & 8             & 0.833                         & 0.725                           & {\color{reda} \textbf{0.029}} & 0.760                         & 0.900                         \\ \midrule[1pt]
 2.1           &                                                                      & 1/8           & 0.844                         & 0.698                           & 0.032                         & 0.744                         & 0.907                         \\
 2.2           &                                                                      & 1/4           & {\color{reda} \textbf{0.845}} & 0.700                           & 0.032                         & 0.746                         & 0.911                         \\
 2.3           &                                                                      & 1/2           & 0.843                         & 0.701                           & 0.031                         & 0.746                         & 0.908                         \\
 2.4           &                                                                      & 1             & 0.842                         & 0.713                           & 0.030                         & 0.754                         & 0.908                         \\
 2.5           &                                                                      & 2             & 0.839                         & 0.720                           & 0.030                         & 0.761                         & 0.908                         \\
 2.6           &                                                                      & 4             & 0.839                         & 0.706                           & 0.032                         & 0.752                         & 0.909                         \\
 2.7           & \multirow{-7}{*}{$\Phi_{exp}^{\alpha}(x)=e^{-(\alpha (x-0.5))^{2}}$} & 8             & 0.841                         & 0.698                           & 0.032                         & 0.745                         & 0.910                         \\ \midrule[1pt]
 3             & BCE w/ $\omega = 1+\Phi_{pow}^{2}(x)$                                & 2             & 0.844                         & 0.697                           & 0.032                         & 0.744                         & {\color{reda} \textbf{0.913}} \\
 \bottomrule[2pt]
\end{tabular}

%% file: section/appendix.tex
\appendix

\begin{figure*}[ht]
	\centering
	\includegraphics[width=\linewidth]{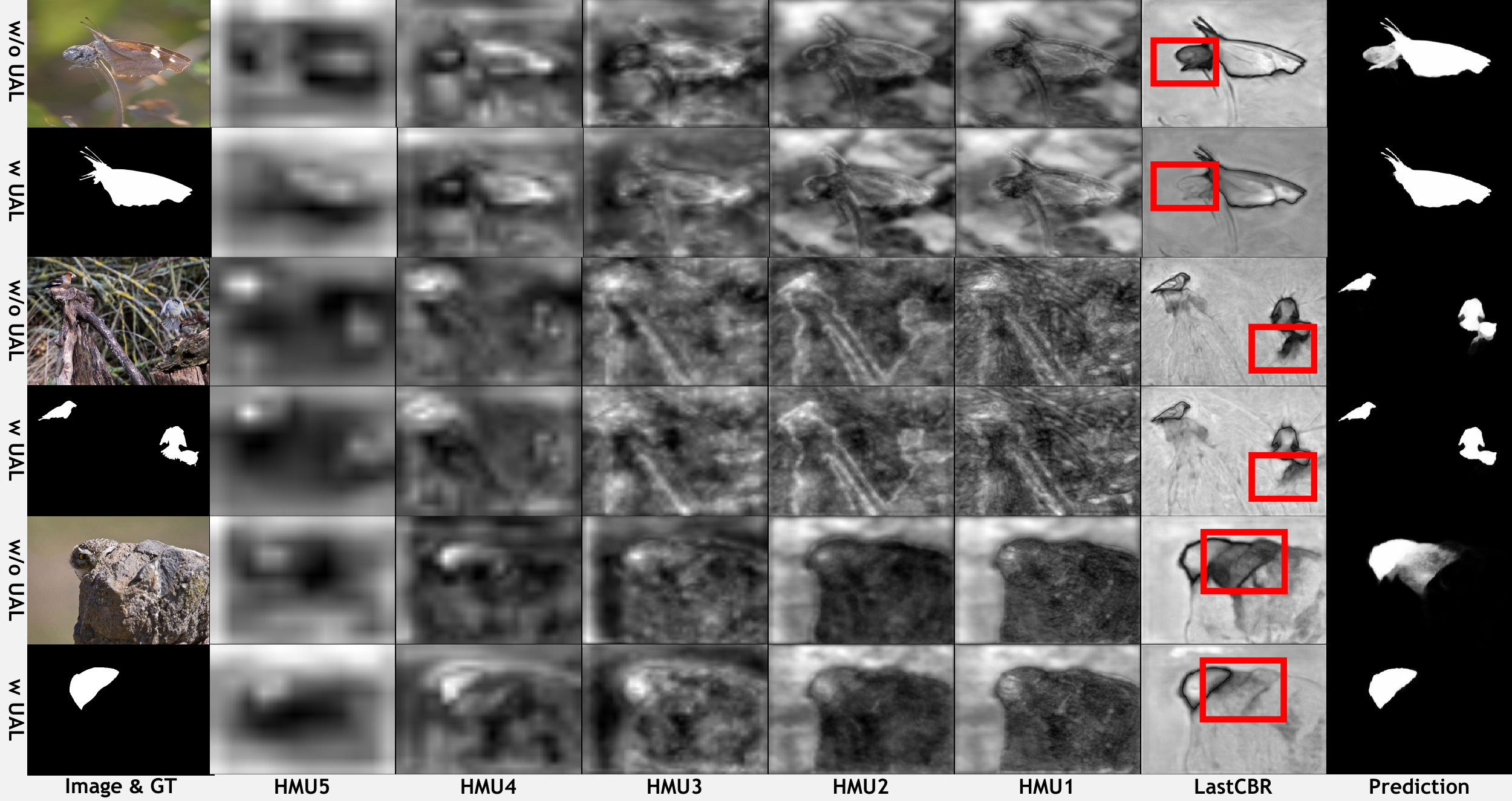}
	\caption{
		Visual comparison of intermediate feature maps from different stages of the decoder for showing the effects of the proposed UAL.
		Please zoom in for more details.
		HMU: Hierarchical mixed-scale unit;
		LastCBR: The last ``Conv$3 \times 3$-BN-ReLU'' structure before the layer generating the logits map.
	}
	\label{fig:featvisualablation}
\end{figure*}

This appendix will introduce more details that cannot be expanded in the main text, while showing the performance on SOD.

\section{Model Details}\label{sec:detailsaboutmodels}

\begin{lstlisting}[
	float=*,
	language=Python,
	firstline=1,
	caption={Code of stacked CBR units.},
	label={listing:stackedcbr},
]
class StackedCBRBlock(nn.Sequential):
    def __init__(self, in_c, out_c, num_blocks=1, kernel_size=3):
        super().__init__()
        self.kernel_setting = dict(kernel_size=kernel_size, stride=1, padding=kernel_size // 2)
        cs = [in_c] + [out_c] * num_blocks
        self.channel_pairs = tuple(self.slide_win_select(cs, win_size=2, win_stride=1, drop_last=True))
        for i, (i_c, o_c) in enumerate(self.channel_pairs):
            self.add_module(name=f"cbr_{i}", module=CBR(i_c, o_c, **self.kernel_setting))
    @staticmethod
    def slide_win_select(items, win_size=1, win_stride=1, drop_last=False):
        i = 0
        while i + win_size <= len(items):
            yield items[i: i + win_size]
            i += win_stride
        if not drop_last:
            yield items[i: i + win_size]
\end{lstlisting}

\subsection{E-Net}\label{sec:e_net}

E-Net is based on the feature extraction part of ResNet-50~\cite{Resnet} and the layers after the ``layer4'' are removed.
We collect the feature maps before passing the first max-pooling layer and the output feature maps of ``layer1'', ``layer2'', ``layer3'' and ``layer4'' as the output feature maps of the E-Net.
The numbers of channels corresponding to them are 64, 256, 512, 1024, and 2048, respectively.

\subsection{C-Net}\label{sec:c_net}

Following the setting of the method~\cite{GateNet}, in C-Net, we use an ASPP~\cite{Deeplab} simplified according to our needs as the feature compression layer corresponding to the ``layer4'' of E-Net and other layers are simply composed of an independent ``Conv$3 \times 3$-BN-ReLU'' ($3 \times 3$ CBR) unit.
The numbers of output channels of all levels are set to $64$ in our models.

The ASPP layer is composed of five CBR branches.
The kernel sizes and dilation rates of them are $1, 3, 3, 3, 1$ and $1, 2, 5, 7, 1$.
All convolution operations use the padding to ensure that the input and output sizes are consistent.
A global average pooling operation and an up-sampling operation are used before and after the second $1 \times 1$ CBR branch to capture the global context information and restore it to the original size.
All results of the five branches are concatenated along the channel dimension and fused by a $3 \times 3$ CBR unit to obtain the output.

\begin{figure}[t]
	\centering
	\includegraphics[width=\linewidth]{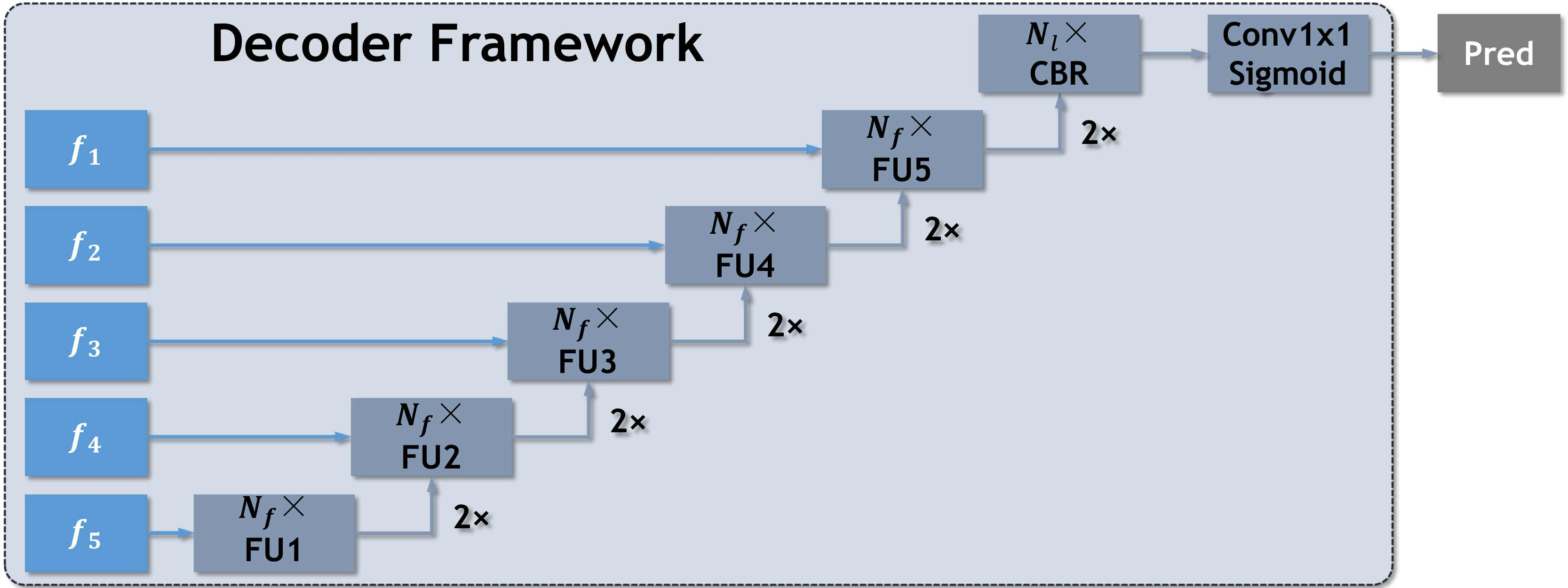}
	\caption{
		Illustration of the basic framework adopted by the decoder in our proposed method.
		FU: The fusion unit for fusing the up-sampled feature map from the previous FU and the shallower feature map $f_i$ ($i=1,2,...,5$).
		$2\times$: The bi-linear interpolation operation with a factor of $2$.
		CBR: The ``Conv$3 \times 3$-BN-ReLU'' unit.
		Conv$1\times1$: The convolution operation with a kernel size of $1 \times 1$.
		$N_f$ and $N_l$: The numbers of FUs and the last CBR units.
	}
	\label{fig:decoder_framework}
\end{figure}

\subsection{Decoder Framework}\label{sec:decoderframework}

The decoder networks of our models in all experiments follow the same framework as shown in Fig.~\ref{fig:decoder_framework}.
Before being fed into the fusion unit (FU), the up-sampled deeper feature map is directly added to the shallow feature map.

In our all experiments, $N_f$ and $N_l$ are set to $1$.
The numbers of input \& output channels of the last $3 \times 3$ CBR unit are $64$ and $32$, respectively.
The number of output channels of the ``Conv$1 \times 1$'' is $1$ and a sigmoid layer is cascaded to convert the logits map to the prediction.
In the decoder of the proposed ZoomNet, the FU is set to the HMU and the other layers remain the same.

\subsection{Baseline Model}\label{sec:baselinemodel}

In the ablation study, we introduce a simple encoder-decoder network as our baseline model to evaluate the performance of different proposed components.
It contains a feature extraction network ``E-Net'', a simple multi-level feature compression convolutional network ``C-Net'', and a basic convolutional decoder where the FU is set to the $3 \times 3$ CBR unit.
In the following text, ``CBR$1$-$5$'' are used to refer to these five units.

\subsection{Model \ding{176}}\label{sec:model_ding176}

In Tab. 2 of the main text, based on the baseline model \ding{172}, we construct the model \ding{176} with the similar amount of parameters and FLOPs to \ding{175} to reflect the effectiveness of the method and the rationality of the design.
For increasing the number of parameters and FLOPs, we made the following modifications to the baseline model \ding{172}:
\begin{itemize}[noitemsep, nolistsep]
	\item The number of output channels of all levels of C-Net: $64 \rightarrow 128$.
	\item The number of input/output channels of CBR$1$-$5$ units of the basic convolutional decoder: $64 \rightarrow 128$.
	\item The number of input channels of the last CBR unit: $64 \rightarrow 128$.
	\item The number of CBR units ($N_f$ and $N_l$) of all levels of the basic convolutional decoder: $1 \rightarrow 3$.
	\item The kernel size of the convolution operation in all levels of the basic convolutional decoder: $3 \rightarrow 5$.
\end{itemize}

To facilitate understanding, the corresponding code for the stacked CBR units used here is listed in List.~\ref{listing:stackedcbr}.

\begin{algorithm}
	\caption{The iteration struction in the HMU}
	\label{alg:iterationinhmu}
	\begin{algorithmic}[1]
		\Require
		$\{g_j\}_{j=1}^{G}$: feature groups;
		$G \geq 2$: the number of groups;
		$C = 32$: the number of channels in a single feature group $g_j$;
		$\mathcal{S}$: splitting operation;
		$\mathcal{T}_{C_o \times C_i}$: stacked CBR units with initial input and final output channel numbers of $C_i$ and $C_o$ as listed in List.~\ref{listing:stackedcbr};
		$\mathcal{C}$: concatenation operation;
		\Ensure
		$\{{g'}^2_j\}_{j=1}^{G}$: the feature set for generating the modulation vector $\alpha$;
		$\{{g'}^3_j\}_{j=1}^{G}$: the feature set used to be modulated and generate the final output of the HMU;
		\For{$i\gets 1, G$}
		\If{$i = 1$} \Comment{Group $1$}
		\State ${g'}^1_{i}, {g'}^2_{i}, {g'}^3_{i} \gets \mathcal{S}(\mathcal{T}^{i}_{3C \times C}(g_i))$;
		\State ${g'}^1_{prev} \gets {g'}^1_{i}$;
		\ElsIf{$i = G$} \Comment{Group $G$}
		\State ${g'}^2_{i}, {g'}^3_{i} \gets \mathcal{S}(\mathcal{T}^{i}_{2C \times 2C}(\mathcal{C}(g_i, {g'}^1_{prev})))$;
		\Else \Comment{Group $i$, $1 < i < G$}
		\State ${g'}^1_{i}, {g'}^2_{i}, {g'}^3_{i} \gets \mathcal{S}(\mathcal{T}^{i}_{3C \times 2C}(\mathcal{C}(g_i, {g'}^1_{prev})))$;
		\State ${g'}^1_{prev} \gets {g'}^1_{i}$;
		\EndIf
		\EndFor
	\end{algorithmic}
\end{algorithm}

\begin{table}[t]
	\centering
	\caption{
		Comparisons of the number of parameters and FLOPs based on~\url{https://github.com/lartpang/MethodsCmp} corresponding to recent COD methods.
		All evaluations follow the inference settings in the corresponding papers.
	}
	\resizebox{\linewidth}{!}{%
		\input{tables/cod_params_flops.tex}
	}
	\label{tab:cod_params_flops}
\end{table}

\begin{figure}[t]
	\centering
	\includegraphics[width=1 \linewidth]{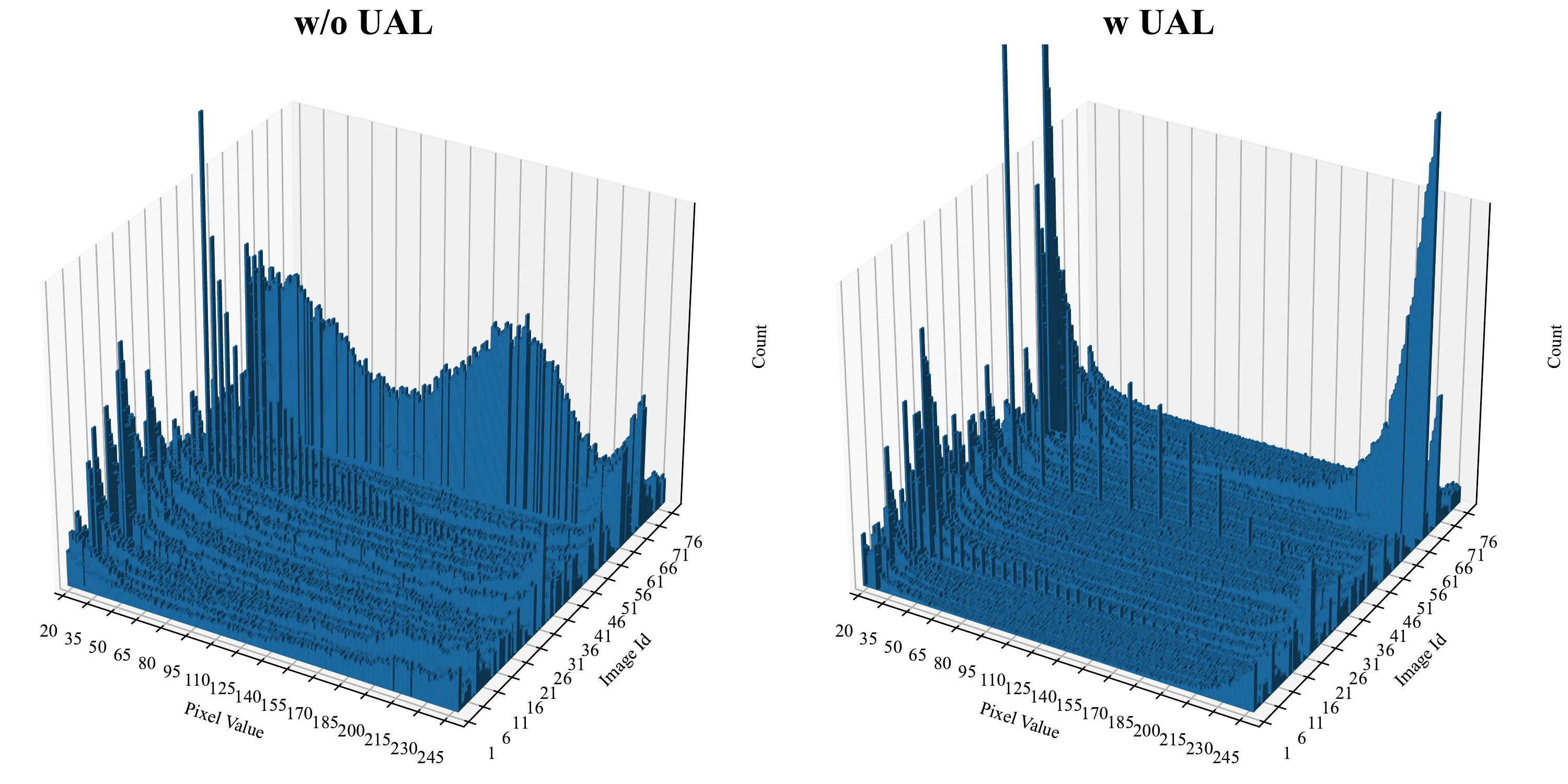}
	\caption{
		Visual comparison of histograms of all 76 prediction results on the CHAMELEON~\cite{CHAMELEON} dataset, which is a stack of the histogram of each prediction.
		A good result should embody a closely binarized histogram at both ends.
		For a more clear demonstration, only the interval with pixel values between 20 and 245 is counted here.
		It is best to zoom in for more details.
	}
	\label{fig:hist}
\end{figure}

\begin{figure*}[t]
	\centering
	\includegraphics[width=0.7\linewidth]{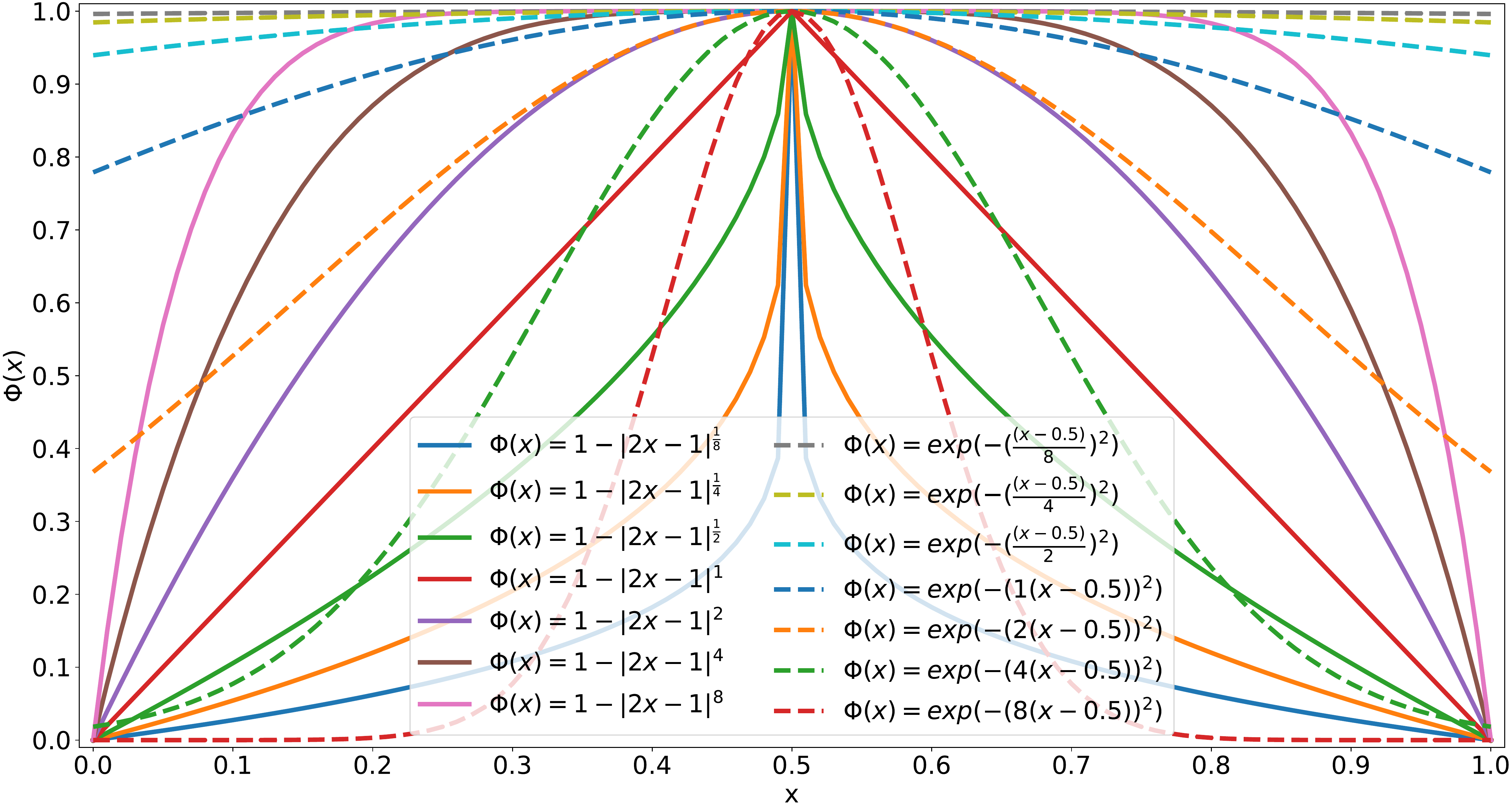}
	\caption{
		Curves of different forms of the proposed UAL.
	}
	\label{fig:loss_form}
\end{figure*}

\begin{table}[t]
	\centering
	\caption{
		Comparisons of different increasing strategies of $\lambda$.
		$\lambda_{const}$: A constant value and it is set to $1$.
		$t$ and $T$: The current and total number of iterations, respectively.
		$\lambda_{min}$ and $\lambda_{max}$: The minimum and maximum values of $\lambda$, and they are set to $0$ and $1$ in our experiments.
		``Linear$_{t_{min} \rightarrow t_{max}}$'': The linearly increasing interval in the iterations is $[t_{min}, t_{max}]$.
		clip: Values outside the interval are clipped to the interval edges.
	}
	\resizebox{\linewidth}{!}{%
		\input{tables/loss_coef.tex}
	}
	\label{tab:exp_losscoef}
\end{table}

\begin{figure*}[t]
	\centering
	\includegraphics[width=\linewidth]{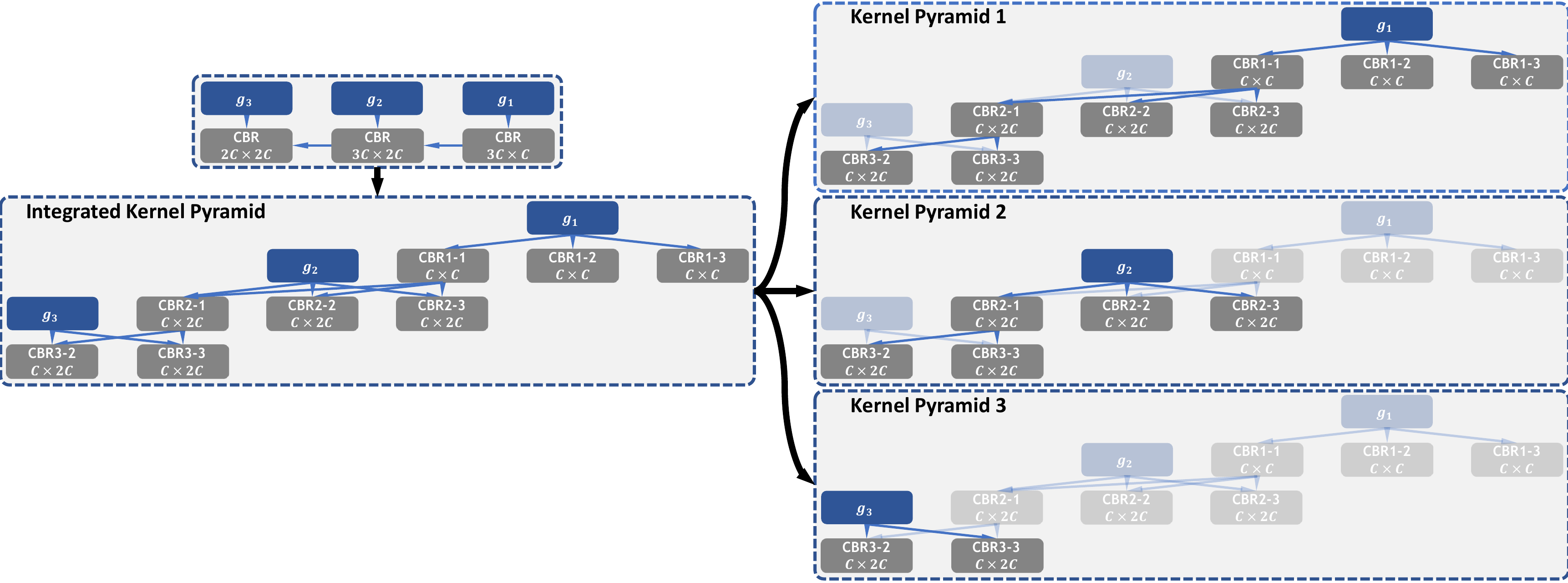}
	\caption{
		The iteration structure of feature groups in HMU can be regarded as an integrated kernel pyramid.
		Without loss of generality, we show the situation of the number of groups $G=3$ in the figure.
		The actual final model is set to $G=6$.
		The only difference lies in the number of repetitions of the kernel pyramid structure in the middle.
		``CBR$l$-$j$'': The ``Conv$3 \times 3$-BN-ReLU'' structure corresponding to the input feature group g$_l$ and the $j$th output feature group.
		$C_o \times C_i$: The numbers of input and output channels of the CBR unit is $C_i$ and $C_o$, respectively.
	}
	\label{fig:kernelpyramid}
\end{figure*}

\section{HMU: Perspective of Kernel Pyramid}\label{sec:perspectiveofkernelpyramid}

The iteration structure of feature groups in HMU is actually equivalent to an integrated multi-path kernel pyramid structure with partial parameter sharing.
In order to understand this intuitively, we highlight the feature information flow of different groups in the iterative structure in Fig.~\ref{fig:kernelpyramid}.
Specifically, the $3 \times 3$ CBR unit corresponding to the feature group in the iteration structure can be split according to the output feature groups.
As shown in the ``Integrated Kernel Pyramid'' on the left of Fig.~\ref{fig:kernelpyramid}, each original CBR unit with an output channel number of $3C$ is converted to three independent CBR units with a shared input.
And the numbers of output channels of them are $C$.
When we further decouple the integrated form on the left into the form on the right, we can clearly see that the information flow paths corresponding to different feature groups each form a multi-branch kernel pyramid structure and there are some shared parameters between these pyramids.

As mentioned in the main text of the paper, some of the channels in the output feature of each branch are used together to generate the modulation vector.
It not only weights the channels inside each branch, but also weights different branches.
If viewed from the aforementioned perspective of the kernel pyramid, such an operation can be seen as a relative modulation of the different kernel pyramids contained in the iterative structure of the HMU.

Besides, in our HMU, $C$ is set to $32$.
The number of channels of the final output feature of the HMU is the same as the input feature, both are 64.
We also list the algorithm of the iteration structure in Alg.~\ref{alg:iterationinhmu} to present the process more clearly and to complement the related statement in the main text.

\begin{figure*}[!h]
	\centering
	\subfloat[PR curves.]{%
		\centering
		\includegraphics[width=\linewidth,height=0.22\linewidth]{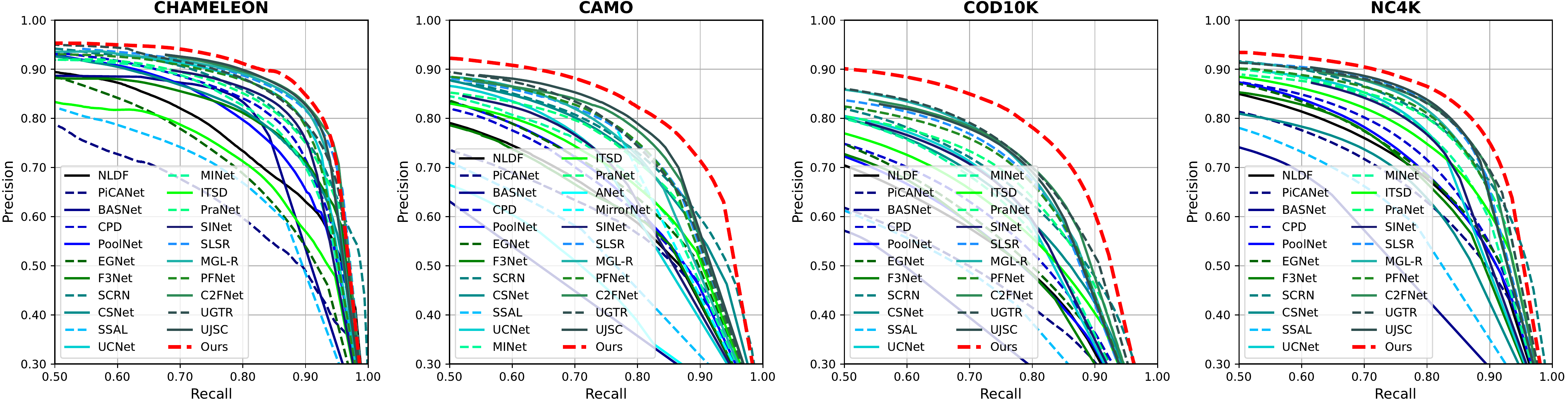}
		\label{fig:pr}
	}
	\\
	\subfloat[$F_{\beta}$ curves.]{%
		\centering
		\includegraphics[width=\linewidth,height=0.22\linewidth]{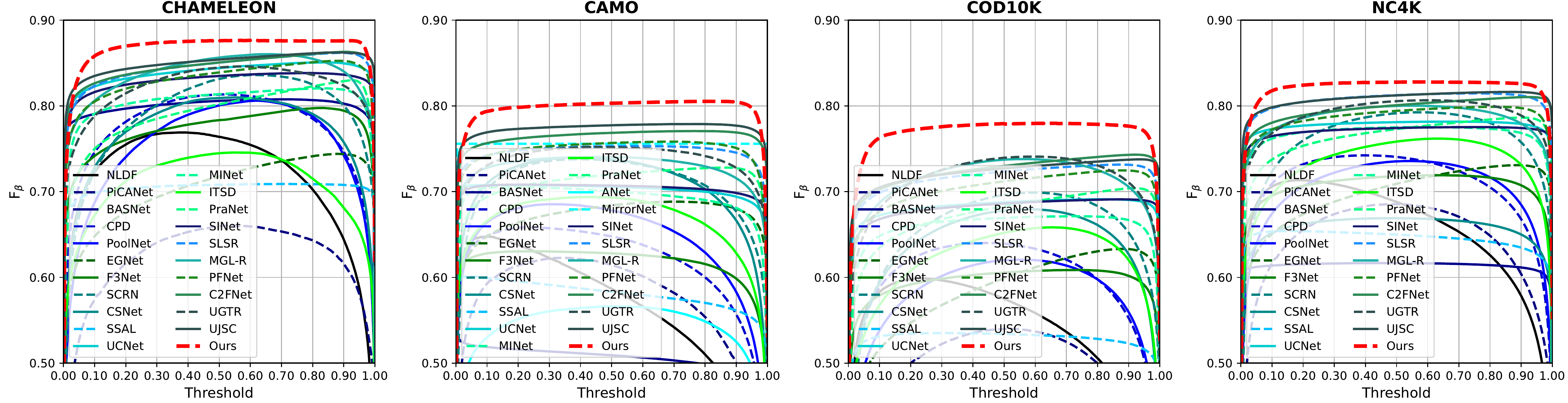}
		\label{fig:fm}
	}
	\caption{PR and $F_{\beta}$ curves of the proposed model and recent SOTA algorithms over four COD datasets.}
	\label{fig:prfm}
\end{figure*}

\section{More Comparisons}

\subsection{PR \& F$_\beta$ curves of COD Methods}\label{sec:cod_curves}

In Fig.~\ref{fig:prfm}, we show the PR \& F$_\beta$ curves of different methods on four COD datasets.
The red curve represents our method.

\subsection{Comparisons of Param. \& FLOPs}\label{sec:params_and_flops}

In Tab.~\ref{tab:cod_params_flops}, we list the number of parameters and FLOPs of existing COD methods and ours.
Our method provides a performance-robust solution with the second-smallest amount of parameters for the COD task.
But there may be still some redundancy in the design of the inference structure.
The adopted explicit scale-independent design may bring additional inference cost.
We will explore and improve this in future work.

\subsection{Intermediate Feature Maps of the Decoder}\label{sec:intermediatefeaturemaps}

We show the intermediate feature maps from different stages of the decoder in Fig.~\ref{fig:featvisualablation}.

\subsection{Effectiveness of UAL}\label{sec:hist}

In Fig.~\ref{fig:hist}, we visualize the histogram maps of all results on CHAMELEON~\cite{CHAMELEON}.

\subsection{Different Forms of $\lambda$}\label{sec:formocoef}

The different adjustment functions of the coefficient $\lambda$ and their results of UAL are list in Tab.~\ref{tab:exp_losscoef}.

\subsection{Different Forms of UAL}\label{sec:formsofual}

The different forms of UAL are shown in Fig.~\ref{fig:loss_form}.

\subsection{Performance in More Complex Scenes}\label{sec:cpd1k}

% \usepackage{graphicx}
\begin{table}
	\centering
	\caption{
		Comparison results of methods trained without CPD1K-TR on CPD1K-TE~\cite{CPD1K}.
	}
	\resizebox{0.7\linewidth}{!}{%
		\input{tables/cpk1k.tex}
	}
	\label{tab:cpd1k}
\end{table}

Actually, COD10K-TE is a very representative test dataset with rich and diverse scenarios and objects.
Besides, there is also a very complex small-scale dataset CPD1K~\cite{CPD1K}.
Tab.~\ref{tab:cpd1k} shows the results of our method and some state-of-the-art competitors (all are trained without CPD1K-TR).
The test results on CPD1K-TE can reflect the adaptability of the model to complex scenarios.
The experiment shows the superior performance of our method in more complex scenarios.

\section{Experiments on SOD}\label{sec:sod}

% \usepackage{graphicx}
\begin{table*}[t]
	\centering
	\caption{
		More detailed comparison results on the SOD task.
		The best results are highlighted in {\color{reda} \textbf{red}}, {\color{mygreen} \textbf{green}} and {\color{myblue} \textbf{blue}}.
		These results are based on the VGG~\cite{VGG}, ResNet~\cite{Resnet} and T2T-ViT~\cite{T2T-ViT} version of the corresponding method.
	}
	\resizebox{\linewidth}{!}{%
		\input{tables/sod_sota.tex}
	}
	\label{tab:detail_totalsod}
\end{table*}

In order to show good generalization and further verify the rationality of the structural design, we evaluate the proposed model on the SOD task.

\subsection{Datasets}\label{sec:datasets}

Our experiment on SOD is based on the existing five SOD datasets,
DUT-OMRON~\cite{DUT-OMRON} (5168),
DUTS~\cite{DUTS} (10553 + 5017),
ECSSD~\cite{ECSSD} (1000),
HKU-IS~\cite{HKU-IS} (4447)
and Pascal-S~\cite{HKU-IS} (850).
We only use the training set of DUTS for training.
During the test phase, we use the remaining data for inference.

\subsection{Implementation Details}\label{sec:implementationdetails}

For a fair comparison on SOD, the proposed model is retrained on DUTS~\cite{DUTS} following the training strategies and techniques of~\cite{SCRN,F3Net,LDF,PFANet,MINet}.
The learning rate is initialized to 0.05 and follows a linear warm-up and linear decay strategy.
And the main scale is changed to $352 \times 352$ to achieve a trade-off between performance and speed.
The model tends to converge after 50 epochs with a batch size of 22.

\begin{figure*}[t]
	\centering
	\subfloat[PR curves.]{%
		\centering
		\includegraphics[width=\linewidth,height=0.2\linewidth]{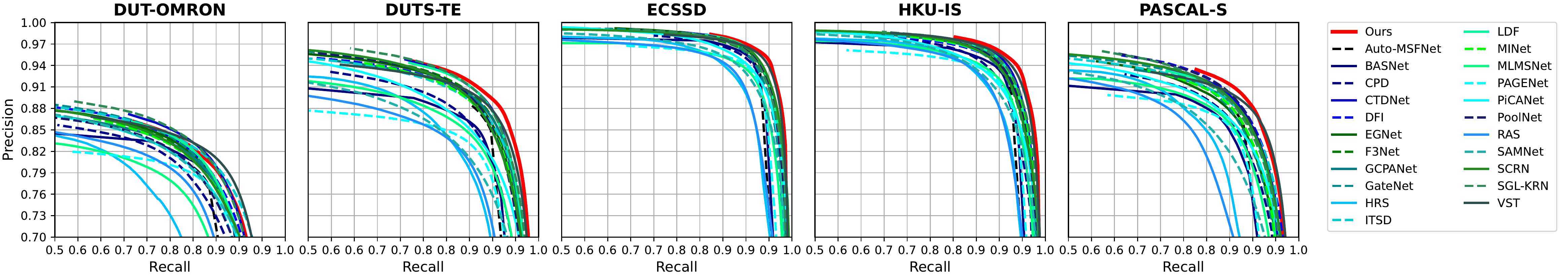}
		\label{fig:sod_pr}
	}
	\\
	\subfloat[$F_{\beta}$ curves.]{%
		\centering
		\includegraphics[width=\linewidth,height=0.2\linewidth]{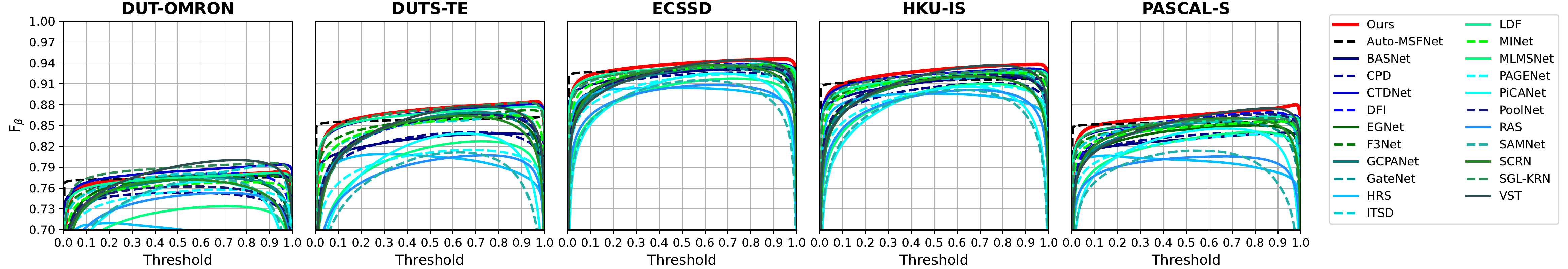}
		\label{fig:sod_fm}
	}
	\caption{PR and $F_{\beta}$ curves of the proposed model and recent SOTA algorithms over five SOD datasets.}
	\label{fig:sod_prfm}
\end{figure*}

\subsection{Comparisons with State-of-the-arts}\label{sec:comp_sota}

We compare the proposed model with 22 existing methods.
All the results are listed in Tab.~\ref{tab:detail_totalsod} and shown in Fig.~\ref{fig:sod_prfm}.
Our model outperforms all these competitors, which shows that the proposed model can deal with the more general binary segmentation task.

\section{Limitations and Future Work}\label{sec:limitationsandfuturework}

Although our ZoomNet provides a powerful and effective solution for the COD task, some limitations still exist and are worth exploring further.

\begin{enumerate}
	\item In the current work, the shared feature extraction structure explicitly collects complementary information from different scales on the image pyramid, which is designed to mimic the behavior of zooming in and out. But for human beings, the process of information extraction and integration should be implicit and internalized in the process of knowledge learning. Moreover, this explicit scale-independent design also brings the additional inference cost. Although our method has achieved good performance on COD and SOD tasks, the inference speed is still slightly slower than the current fastest method, C2FNet~\cite{COD-C2FNet}.
	\item Besides, there is still room for improvement in the way of mining effective clues from small-scale features in SIU.
\end{enumerate}

In future work, we will try to further simplify the inference structure to make it more in line with the actual human decision-making process and optimize the ability of our method to extract contextual cues from small-scale features.

%% file: tables/cod_params_flops.tex
\begin{tabular}{c|cccccccc}
 \toprule[2pt]
 Method  & Ours     & UGTR~\cite{COD-UGTR} & C$^{2}$F-Net~\cite{COD-C2FNet} & UJSC~\cite{UJSC} & PFNet~\cite{COD-PFNet} & MGL-R~\cite{COD-MGL} & SLSR~\cite{SLSR} & SINet~\cite{COD10K} \\ \midrule[1pt]
 Params. & 32.382M  & 48.868M              & 28.411M                        & 217.982M         & 46.498M                & 63.595M              & 50.935M          & 48.947M             \\
 FLOPs   & 203.496G & 1.007T               & 26.167G                        & 112.341G         & 53.222G                & 553.939G             & 66.625G          & 38.757              \\
 FPS     & 24.030   & 16.640               & 65.759                         & 34.178           & 62.590                 & 13.373               & 58.782           & 56.509              \\
 \bottomrule[2pt]
\end{tabular}

%% file: tables/loss_coef.tex
\begin{tabular}{c|c|ccccc}
 \toprule[2pt]
 \rowcolor{tabtitle}
 Strategy                         & $\lambda$                                                                                                                                           & S$_{m}$ $\uparrow$ & F$^{\omega}_{\beta}$ $\uparrow$ & MAE $\downarrow$ & F$_{\beta}$ $\uparrow$ & E$_{m}$ $\uparrow$ \\ \midrule[1pt] \rowcolor{ours}
 Cosine                           & $\lambda_{min} + \frac{1}{2}(1 - \cos(\frac{t}{T}\pi))(\lambda_{max} - \lambda_{min})$                                                              & 0.838              & 0.729                           & 0.029            & 0.766                  & 0.911              \\ \midrule[1pt]
 Linear$_{0 \rightarrow T}$       &                                                                                                                                                     & 0.834              & 0.723                           & 0.029            & 0.760                  & 0.908              \\
 Linear$_{0.3T \rightarrow 0.7T}$ & \multirow{-2}{*}{$\text{clip}(\lambda_{min} + \frac{t - t_{min}}{t_{max} - t_{min}}(\lambda_{max} - \lambda_{min}), \lambda_{min}, \lambda_{max})$} & 0.832              & 0.719                           & 0.030            & 0.758                  & 0.904              \\ \midrule[1pt]
 Constant                         & $\lambda_{const}$                                                                                                                                   & 0.830              & 0.717                           & 0.030            & 0.757                  & 0.906              \\
 \bottomrule[2pt]
\end{tabular}%

%% file: tables/cpk1k.tex
\begin{tabular}{c|ccccc}
 \toprule[2pt]
 \rowcolor{tabtitle}
 Model     & S$_{m}$ $\uparrow$ & F$^{\omega}_{\beta}$ $\uparrow$ & MAE $\downarrow$ & F$_{\beta}$ $\uparrow$ & E$_{m}$ $\uparrow$ \\
 \midrule[1pt]
 ZoomNet   & \textbf{0.759}     & \textbf{0.537}                  & \textbf{0.011}   & \textbf{0.578}         & \textbf{0.843}     \\
 C$^2$FNet & 0.743              & 0.495                           & 0.016            & 0.528                  & 0.840              \\
 PFNet     & 0.722              & 0.460                           & 0.017            & 0.494                  & 0.819              \\
 \bottomrule[2pt]
\end{tabular}%

%% file: tables/sod_sota.tex
\begin{tabular}{l|c|c*{5}{*{5}{|c}}}
 \toprule[2pt] \rowcolor{tabtitle}
                                &                            &                        & \multicolumn{5}{c}{\textbf{DUT-OMRON}} & \multicolumn{5}{|c}{\textbf{DUTS-TE}} & \multicolumn{5}{|c}{\textbf{ECSSD}} & \multicolumn{5}{|c}{\textbf{HKU-IS}} & \multicolumn{5}{|c}{\textbf{PASCAL-S}}                                                                                                                                                                                                                                                                                                                                                                                                                                                                                                                                                                                                                                                                                                                             \\ \rowcolor{tabtitle}
 \multirow{-2}{*}{Model}        & \multirow{-2}{*}{Backbone} & \multirow{-2}{*}{Year} & S$_{m}$ $\uparrow$                     & F$^{\omega}_{\beta}$ $\uparrow$       & MAE $\downarrow$                    & F$_{\beta}$ $\uparrow$               & E$_{m}$ $\uparrow$                     & S$_{m}$ $\uparrow$               & F$^{\omega}_{\beta}$ $\uparrow$  & MAE $\downarrow$                 & F$_{\beta}$ $\uparrow$           & E$_{m}$ $\uparrow$               & S$_{m}$ $\uparrow$               & F$^{\omega}_{\beta}$ $\uparrow$  & MAE $\downarrow$                 & F$_{\beta}$ $\uparrow$           & E$_{m}$ $\uparrow$               & S$_{m}$ $\uparrow$               & F$^{\omega}_{\beta}$ $\uparrow$  & MAE $\downarrow$                 & F$_{\beta}$ $\uparrow$           & E$_{m}$ $\uparrow$               & S$_{m}$ $\uparrow$               & F$^{\omega}_{\beta}$ $\uparrow$  & MAE $\downarrow$                 & F$_{\beta}$ $\uparrow$           & E$_{m}$ $\uparrow$               \\ \midrule[1pt]
 RAS~\cite{RAS}                 & VGG16                      & 2018                   & 0.814                                  & 0.695                                 & 0.062                               & 0.731                                & 0.860                                  & 0.839                            & 0.740                            & 0.059                            & 0.779                            & 0.889                            & 0.893                            & 0.857                            & 0.056                            & 0.887                            & 0.931                            & 0.887                            & 0.843                            & 0.045                            & 0.875                            & 0.940                            & 0.793                            & 0.735                            & 0.106                            & 0.790                            & 0.846                            \\
 MLMSNet~\cite{MLMSNet}         & VGG16                      & 2019                   & 0.809                                  & 0.681                                 & 0.064                               & 0.710                                & 0.848                                  & 0.862                            & 0.761                            & 0.049                            & 0.792                            & 0.907                            & 0.911                            & 0.871                            & 0.045                            & 0.890                            & 0.944                            & 0.907                            & 0.859                            & 0.039                            & 0.878                            & 0.950                            & 0.845                            & 0.785                            & 0.075                            & 0.814                            & 0.893                            \\
 PAGENet~\cite{PAGE-Net}        & VGG16                      & 2019                   & 0.824                                  & 0.722                                 & 0.062                               & 0.743                                & 0.858                                  & 0.854                            & 0.769                            & 0.052                            & 0.793                            & 0.896                            & 0.912                            & 0.886                            & 0.042                            & 0.904                            & 0.947                            & 0.903                            & 0.865                            & 0.037                            & 0.884                            & 0.948                            & 0.838                            & 0.789                            & 0.079                            & 0.819                            & 0.885                            \\
 PiCANet~\cite{PiCANet}         & ResNet50                   & 2018                   & 0.832                                  & 0.695                                 & 0.065                               & 0.729                                & 0.876                                  & 0.869                            & 0.755                            & 0.051                            & 0.791                            & 0.920                            & 0.917                            & 0.867                            & 0.046                            & 0.890                            & 0.952                            & 0.904                            & 0.840                            & 0.043                            & 0.866                            & 0.950                            & 0.852                            & 0.779                            & 0.078                            & 0.812                            & 0.899                            \\
 BASNet~\cite{BASNet}           & ResNet34                   & 2019                   & 0.836                                  & 0.751                                 & 0.056                               & 0.767                                & 0.871                                  & 0.866                            & 0.803                            & 0.048                            & 0.823                            & 0.903                            & 0.916                            & 0.904                            & 0.037                            & 0.917                            & 0.951                            & 0.909                            & 0.889                            & 0.032                            & 0.902                            & 0.951                            & 0.834                            & 0.797                            & 0.079                            & 0.824                            & 0.883                            \\
 CPD~\cite{CPD}                 & ResNet50                   & 2019                   & 0.825                                  & 0.719                                 & 0.056                               & 0.742                                & 0.868                                  & 0.869                            & 0.795                            & 0.043                            & 0.821                            & 0.914                            & 0.918                            & 0.898                            & 0.037                            & 0.913                            & 0.951                            & 0.905                            & 0.875                            & 0.034                            & 0.892                            & 0.950                            & 0.844                            & 0.800                            & 0.074                            & 0.827                            & 0.888                            \\
 PoolNet~\cite{PoolNet}         & ResNet50                   & 2019                   & 0.831                                  & 0.725                                 & 0.054                               & 0.747                                & 0.867                                  & 0.887                            & 0.817                            & 0.037                            & 0.840                            & 0.926                            & 0.926                            & 0.904                            & 0.035                            & 0.918                            & 0.956                            & 0.919                            & 0.888                            & 0.030                            & 0.903                            & 0.958                            & 0.864                            & 0.819                            & 0.067                            & 0.846                            & 0.905                            \\
 EGNet~\cite{EGNet}             & ResNet50                   & 2019                   & 0.841                                  & 0.738                                 & 0.053                               & 0.760                                & 0.878                                  & 0.887                            & 0.816                            & 0.039                            & 0.839                            & 0.927                            & 0.925                            & 0.903                            & 0.037                            & 0.918                            & 0.955                            & 0.918                            & 0.887                            & 0.031                            & 0.902                            & 0.958                            & 0.850                            & 0.804                            & 0.076                            & 0.833                            & 0.892                            \\
 HRS~\cite{HRS}                 & ResNet50                   & 2019                   & 0.772                                  & 0.645                                 & 0.066                               & 0.690                                & 0.841                                  & 0.829                            & 0.746                            & 0.051                            & 0.791                            & 0.899                            & 0.883                            & 0.859                            & 0.054                            & 0.894                            & 0.934                            & 0.882                            & 0.851                            & 0.042                            & 0.883                            & 0.941                            & 0.799                            & 0.744                            & 0.091                            & 0.792                            & 0.866                            \\
 SCRN~\cite{SCRN}               & ResNet50                   & 2019                   & 0.837                                  & 0.720                                 & 0.056                               & 0.749                                & 0.875                                  & 0.885                            & 0.803                            & 0.040                            & 0.833                            & 0.925                            & {\color{myblue} \textbf{0.927}}  & 0.900                            & 0.037                            & 0.916                            & 0.956                            & 0.916                            & 0.876                            & 0.034                            & 0.894                            & 0.956                            & {\color{myblue} \textbf{0.865}}  & 0.813                            & 0.066                            & 0.840                            & 0.906                            \\
 F3Net~\cite{F3Net}             & ResNet50                   & 2020                   & 0.838                                  & 0.747                                 & 0.053                               & 0.766                                & 0.872                                  & 0.888                            & 0.835                            & {\color{myblue} \textbf{0.035}}  & 0.852                            & 0.927                            & 0.924                            & 0.912                            & {\color{myblue} \textbf{0.033}}  & {\color{myblue} \textbf{0.925}}  & 0.955                            & 0.917                            & 0.900                            & {\color{myblue} \textbf{0.028}}  & 0.910                            & 0.958                            & 0.857                            & 0.823                            & 0.064                            & 0.843                            & 0.901                            \\
 GCPANet~\cite{GCPANet}         & ResNet50                   & 2020                   & 0.839                                  & 0.734                                 & 0.056                               & 0.756                                & 0.869                                  & 0.891                            & 0.821                            & 0.038                            & 0.841                            & 0.929                            & {\color{myblue} \textbf{0.927}}  & 0.903                            & 0.035                            & 0.916                            & 0.955                            & 0.920                            & 0.889                            & 0.031                            & 0.901                            & 0.958                            & 0.864                            & 0.819                            & {\color{myblue} \textbf{0.063}}  & 0.840                            & 0.906                            \\
 LDF~\cite{LDF}                 & ResNet50                   & 2020                   & 0.839                                  & 0.752                                 & {\color{myblue} \textbf{0.052}}     & 0.770                                & 0.869                                  & 0.892                            & {\color{myblue} \textbf{0.845}}  & {\color{mygreen} \textbf{0.034}} & 0.861                            & 0.930                            & 0.924                            & {\color{myblue} \textbf{0.915}}  & 0.034                            & {\color{mygreen} \textbf{0.927}} & 0.954                            & 0.919                            & {\color{myblue} \textbf{0.904}}  & {\color{myblue} \textbf{0.028}}  & 0.913                            & 0.958                            & 0.859                            & {\color{myblue} \textbf{0.829}}  & {\color{mygreen} \textbf{0.062}} & {\color{myblue} \textbf{0.851}}  & 0.905                            \\
 DFI~\cite{DFI}                 & ResNet50                   & 2020                   & 0.840                                  & 0.738                                 & 0.055                               & 0.762                                & 0.877                                  & 0.887                            & 0.817                            & 0.039                            & 0.840                            & 0.928                            & {\color{myblue} \textbf{0.927}}  & 0.906                            & 0.035                            & 0.920                            & 0.957                            & 0.919                            & 0.890                            & 0.031                            & 0.903                            & {\color{myblue} \textbf{0.961}}  & 0.864                            & 0.824                            & 0.066                            & 0.849                            & 0.907                            \\
 GateNet~\cite{GateNet}         & ResNet50                   & 2020                   & 0.838                                  & 0.729                                 & 0.055                               & 0.757                                & 0.876                                  & 0.885                            & 0.809                            & 0.040                            & 0.837                            & 0.928                            & 0.920                            & 0.894                            & 0.040                            & 0.913                            & 0.952                            & 0.915                            & 0.880                            & 0.033                            & 0.897                            & 0.955                            & 0.854                            & 0.804                            & 0.071                            & 0.835                            & 0.900                            \\
 ITSD~\cite{ITSD}               & ResNet50                   & 2020                   & 0.840                                  & 0.750                                 & 0.061                               & 0.768                                & 0.880                                  & 0.885                            & 0.824                            & 0.041                            & 0.840                            & 0.929                            & 0.925                            & 0.910                            & 0.034                            & 0.921                            & {\color{myblue} \textbf{0.959}}  & 0.917                            & 0.894                            & 0.031                            & 0.904                            & 0.960                            & 0.859                            & 0.823                            & 0.066                            & 0.843                            & {\color{myblue} \textbf{0.910}}  \\
 MINet~\cite{MINet}             & ResNet50                   & 2020                   & 0.833                                  & 0.738                                 & 0.056                               & 0.757                                & 0.869                                  & 0.884                            & 0.825                            & 0.037                            & 0.844                            & 0.927                            & 0.925                            & 0.911                            & {\color{myblue} \textbf{0.033}}  & 0.923                            & 0.957                            & 0.919                            & 0.897                            & 0.029                            & 0.909                            & 0.960                            & 0.854                            & 0.818                            & 0.066                            & 0.841                            & 0.901                            \\
 VST~\cite{VST}                 & T2T-ViT$_t$-14             & 2021                   & {\color{reda} \textbf{0.850}}          & 0.755                                 & 0.058                               & {\color{myblue} \textbf{0.774}}      & {\color{reda} \textbf{0.888}}          & {\color{mygreen} \textbf{0.896}} & 0.828                            & 0.037                            & 0.845                            & {\color{reda} \textbf{0.939}}    & {\color{mygreen} \textbf{0.932}} & 0.910                            & {\color{myblue} \textbf{0.033}}  & 0.920                            & {\color{reda} \textbf{0.964}}    & {\color{mygreen} \textbf{0.928}} & 0.897                            & 0.029                            & 0.907                            & {\color{reda} \textbf{0.968}}    & {\color{reda} \textbf{0.871}}    & 0.827                            & {\color{mygreen} \textbf{0.062}} & 0.847                            & {\color{reda} \textbf{0.918}}    \\
 SAMNet~\cite{SAMNet}           & Handcraft                  & 2021                   & 0.830                                  & 0.699                                 & 0.065                               & 0.734                                & 0.877                                  & 0.849                            & 0.729                            & 0.058                            & 0.768                            & 0.901                            & 0.907                            & 0.858                            & 0.050                            & 0.883                            & 0.945                            & 0.898                            & 0.837                            & 0.045                            & 0.864                            & 0.946                            & 0.822                            & 0.743                            & 0.095                            & 0.784                            & 0.869                            \\
 SGL-KRN~\cite{KRNet}           & ResNet50                   & 2021                   & {\color{mygreen} \textbf{0.846}}       & {\color{reda} \textbf{0.765}}         & {\color{reda} \textbf{0.049}}       & {\color{reda} \textbf{0.783}}        & {\color{mygreen} \textbf{0.885}}       & {\color{myblue} \textbf{0.893}}  & {\color{mygreen} \textbf{0.847}} & {\color{mygreen} \textbf{0.034}} & {\color{mygreen} \textbf{0.865}} & {\color{reda} \textbf{0.939}}    & 0.923                            & 0.910                            & 0.036                            & 0.924                            & 0.954                            & {\color{myblue} \textbf{0.921}}  & {\color{myblue} \textbf{0.904}}  & {\color{myblue} \textbf{0.028}}  & {\color{myblue} \textbf{0.915}}  & {\color{myblue} \textbf{0.961}}  & 0.854                            & 0.823                            & 0.070                            & 0.849                            & 0.900                            \\
 CTDNet~\cite{CTDNet}           & ResNet50                   & 2021                   & {\color{myblue} \textbf{0.844}}        & {\color{mygreen} \textbf{0.762}}      & {\color{myblue} \textbf{0.052}}     & {\color{mygreen} \textbf{0.779}}     & {\color{myblue} \textbf{0.881}}        & {\color{myblue} \textbf{0.893}}  & {\color{mygreen} \textbf{0.847}} & {\color{mygreen} \textbf{0.034}} & {\color{myblue} \textbf{0.862}}  & {\color{myblue} \textbf{0.935}}  & 0.925                            & {\color{myblue} \textbf{0.915}}  & {\color{mygreen} \textbf{0.032}} & {\color{mygreen} \textbf{0.927}} & 0.956                            & {\color{myblue} \textbf{0.921}}  & {\color{mygreen} \textbf{0.909}} & {\color{mygreen} \textbf{0.027}} & {\color{mygreen} \textbf{0.918}} & {\color{myblue} \textbf{0.961}}  & 0.859                            & {\color{myblue} \textbf{0.829}}  & 0.064                            & {\color{myblue} \textbf{0.851}}  & 0.904                            \\
 Auto-MSFNet~\cite{Auto-MSFNet} & ResNet50                   & 2021                   & 0.832                                  & {\color{myblue} \textbf{0.757}}       & {\color{mygreen} \textbf{0.050}}    & 0.772                                & 0.875                                  & 0.877                            & 0.841                            & {\color{mygreen} \textbf{0.034}} & 0.855                            & 0.931                            & 0.914                            & {\color{mygreen} \textbf{0.916}} & {\color{myblue} \textbf{0.033}}  & {\color{mygreen} \textbf{0.927}} & 0.954                            & 0.908                            & 0.903                            & {\color{mygreen} \textbf{0.027}} & 0.912                            & 0.959                            & 0.849                            & {\color{mygreen} \textbf{0.830}} & {\color{myblue} \textbf{0.063}}  & {\color{mygreen} \textbf{0.852}} & 0.902                            \\ \rowcolor{ours}
 Ours                           & ResNet50                   & 2021                   & 0.841                                  & 0.755                                 & 0.053                               & 0.771                                & 0.872                                  & {\color{reda} \textbf{0.900}}    & {\color{reda} \textbf{0.854}}    & {\color{reda} \textbf{0.033}}    & {\color{reda} \textbf{0.866}}    & {\color{mygreen} \textbf{0.936}} & {\color{reda} \textbf{0.935}}    & {\color{reda} \textbf{0.926}}    & {\color{reda} \textbf{0.027}}    & {\color{reda} \textbf{0.933}}    & {\color{mygreen} \textbf{0.963}} & {\color{reda} \textbf{0.931}}    & {\color{reda} \textbf{0.918}}    & {\color{reda} \textbf{0.023}}    & {\color{reda} \textbf{0.923}}    & {\color{mygreen} \textbf{0.967}} & {\color{mygreen} \textbf{0.869}} & {\color{reda} \textbf{0.844}}    & {\color{reda} \textbf{0.057}}    & {\color{reda} \textbf{0.860}}    & {\color{mygreen} \textbf{0.917}} \\ \bottomrule[1pt]
\end{tabular}%